\documentclass{article} 
\usepackage{colm2025_conference}

\usepackage{microtype}
\usepackage{hyperref}
\usepackage{url}
\usepackage{booktabs}
\usepackage{multirow}
\usepackage[table]{xcolor}

\usepackage{lineno}
\usepackage{xcolor}
\usepackage{titlesec}
\usepackage[most]{tcolorbox}
\usepackage{caption}
\usepackage{xspace}

\definecolor{darkblue}{rgb}{0, 0, 0.5}
\hypersetup{colorlinks=true, citecolor=darkblue, linkcolor=darkblue, urlcolor=darkblue}
\usepackage{enumitem}
\usepackage{graphicx}
\usepackage{subcaption}
\usepackage{colortbl}
\usepackage{amsmath}
\usepackage[capitalise,noabbrev]{cleveref}
\usepackage{mathtools}
\usepackage{siunitx}    
\usepackage{makecell,threeparttable}
\crefname{section}{Section}{Sections}
\Crefname{section}{Section}{Sections}
\crefrangelabelformat{section}{#3#1--#4#2}
\definecolor{richpurple}{RGB}{75,46,131}
\newcommand{\purplecomic}[1]{%
  {\color{richpurple}\selectfont #1}%
}

\newcommand{\blackcomic}[1]{%
  {\color{black}\selectfont #1}%
}
\renewcommand{\thefootnote}{\fnsymbol{footnote}}

\newcommand{\AgentEvolver}{\texttt{AgentEvolver}\xspace}
\newcommand\eat[1]{}

\title{
  \raisebox{-2.5 mm}{\includegraphics[width=0.055\textwidth]{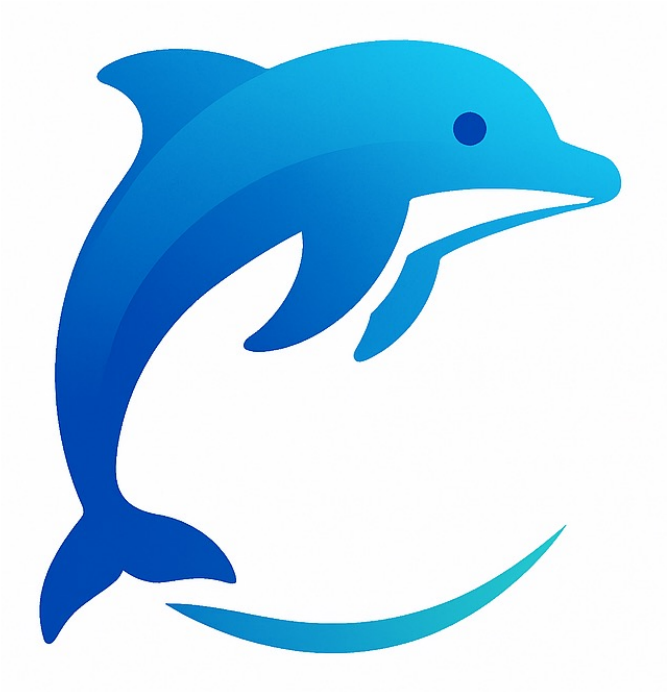}}\hspace{0.2em}\purplecomic{\textbf{\AgentEvolver}}: 
  \blackcomic{Towards Efficient Self-Evolving Agent System}
}

\author{
Yunpeng Zhai\footnotemark[1],
~Shuchang Tao\footnotemark[1], 
~Cheng Chen\footnotemark[1], 
~Anni Zou\footnotemark[1], 
~Ziqian Chen\footnotemark[1], 
~Qingxu Fu\footnotemark[1], \\
~Shinji Mai\footnotemark[1], 
~Li Yu, 
~Jiaji Deng, 
~Zouying Cao, 
~Zhaoyang Liu\footnotemark[2], 
~Bolin Ding\footnotemark[2], 
~Jingren Zhou \\
[1em]
Tongyi Lab\includegraphics[height=12pt]{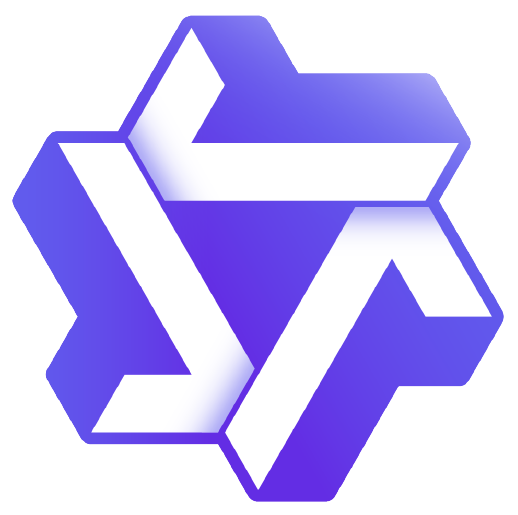}, Alibaba Group \\
[1em]
\texttt{\{zhaiyunpeng.zyp, taoshuchang.tsc, chengchen, zouanni.zan, eric.czq, fuqingxu.fqx, jingmu.lzy, bolin.ding\}@alibaba-inc.com}
}

\begin{document}


\maketitle

\begingroup
  \renewcommand\thefootnote{}%
  \footnotetext{%
    \begin{tabular}{@{}l@{\hspace{0.4em}}l@{}}
      $^{*}$ & Equal contribution.\\
      $^{\dagger}$ & Corresponding authors.
    \end{tabular}%
  }%
\endgroup

\begin{abstract}

Autonomous agents powered by large language models (LLMs) have the potential to significantly enhance human productivity by reasoning, using tools, and executing complex tasks in diverse environments. However, current approaches to developing such agents remain costly and inefficient, as they typically require manually constructed task datasets and reinforcement learning (RL) pipelines with extensive random exploration. These limitations lead to prohibitively high data-construction costs, low exploration efficiency, and poor sample utilization. To address these challenges, we present \textbf{\AgentEvolver}, a self-evolving agent system that leverages the semantic understanding and reasoning capabilities of LLMs to drive autonomous agent learning. \AgentEvolver introduces three synergistic mechanisms: (i) \texttt{self-questioning}, which enables curiosity-driven task generation in novel environments, reducing dependence on handcrafted datasets; (ii) \texttt{self-navigating}, which improves exploration efficiency through experience reuse and hybrid policy guidance; and (iii) \texttt{self-attributing}, which enhances sample efficiency by assigning differentiated rewards to trajectory states and actions based on their contribution. 
By integrating these mechanisms into a unified framework, \AgentEvolver enables scalable, cost-effective, and continual improvement of agent capabilities. Preliminary experiments indicate that \AgentEvolver achieves more efficient exploration, better sample utilization, and faster adaptation compared to traditional RL-based baselines.

\end{abstract}

\begin{figure}[htbp]
  \centering
  \vspace{0em}
  \includegraphics[width=0.95\textwidth]{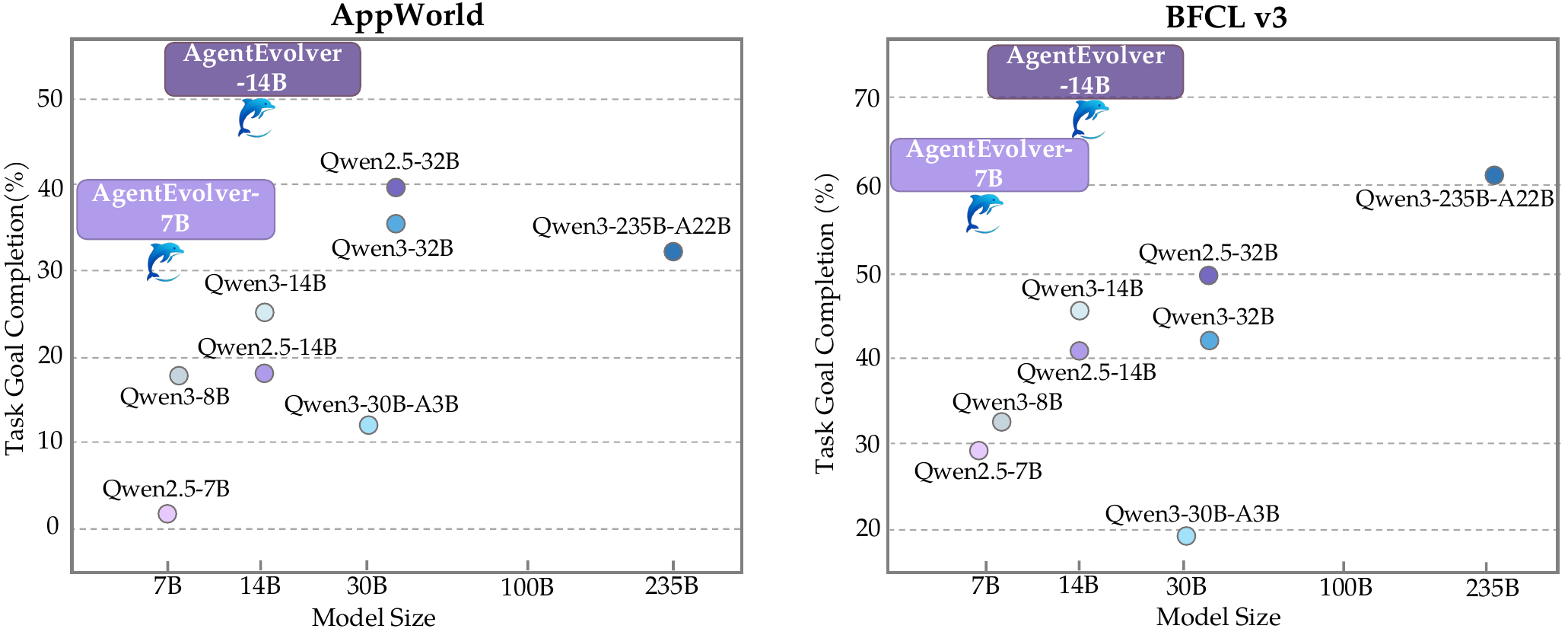}
  \captionsetup{width=0.95\textwidth}
  \caption{Performance comparison on the AppWorld and BFCL-v3 benchmarks. \AgentEvolver achieves superior results while using substantially fewer parameters than larger baseline models.}
  \label{fig:plot_comparison}
\end{figure}

\newpage

\section{Introduction}

The rapid progress of artificial intelligence (AI) and large language models (LLMs) ~\citep{liu2024deepseek, qwen2025technical} has opened new opportunities for developing autonomous agents capable of tool use, reasoning, and task execution in complex environments. Such agents promise to significantly enhance human productivity by automatically interacting with digital systems and real-world platforms~\citep{DBLP:conf/iclr/YaoZYDSN023, DBLP:conf/acl/DingTPWGDSC25,DBLP:journals/fcsc/QuDWCWYXW25}. 
Broadly, the development of agent systems follows two paradigms. One relies on workflow-based orchestration, where developers manually design action sequences or tool-calling rules~\citep{DBLP:conf/icml/DuW024, DBLP:conf/acl/0001YTSLD025}. The other is learning-based optimization, where agents adapt their policies from feedback and interaction~\citep{DBLP:journals/corr/abs-2503-23383,DBLP:journals/corr/abs-2509-13761}. Among learning-based approaches, reinforcement learning (RL) has emerged as a central direction~\citep{DBLP:journals/corr/abs-2509-12867}, offering a principled way to endow agents with adaptive, generalizable behaviors.

Despite this promise, RL-driven agent development remains challenging in practice, especially when adapting to novel environments~\citep{wang2025modeling,DBLP:journals/corr/abs-2509-02547}. 
First, the construction of training tasks tailored for RL is prohibitively expensive. In novel environments, tool functionalities are often unknown, and manually creating tasks with sufficiently diverse trajectories is labor-intensive. 
Second, existing RL methods for LLM-based agents suffer from severe sample inefficiency. Although the RL community has proposed various exploration strategies such as intrinsic rewards~\citep{DBLP:journals/corr/abs-2502-01456} and UCB-based bandit methods~\citep{song2025outcome}, these techniques are often ineffective for long-horizon, tool-augmented agents, where each rollout is computationally and financially expensive. As a result, current LLM agent training still primarily adopts PPO- or GRPO-style policy optimization~\citep{deepseek_math}, whose reliance on massive trajectory sampling leads to approximate brute-force exploration with many redundant rollouts and limited learning value. Consequently, the training of LLM agents remains a laborious and costly paradigm.

Motivated by the rapidly growing semantic understanding and reasoning capabilities of LLMs, a natural question arises: \textbf{why not entrust the model itself with greater autonomy in driving its own learning process?} Instead of relying on rigid, human-engineered pipelines, we envision an agent system where the LLM actively guides exploration, task generation, and performance refinement. To this end, we propose \textbf{\AgentEvolver}, a self-evolving agent system designed to achieve autonomous and efficient capability evolution through environmental interaction. By harnessing the strengths of LLMs, \AgentEvolver establishes a self-training loop that continuously improves agent competence through direct environment interaction.

\begin{figure}
  \centering
  \includegraphics[width=1.0\textwidth]{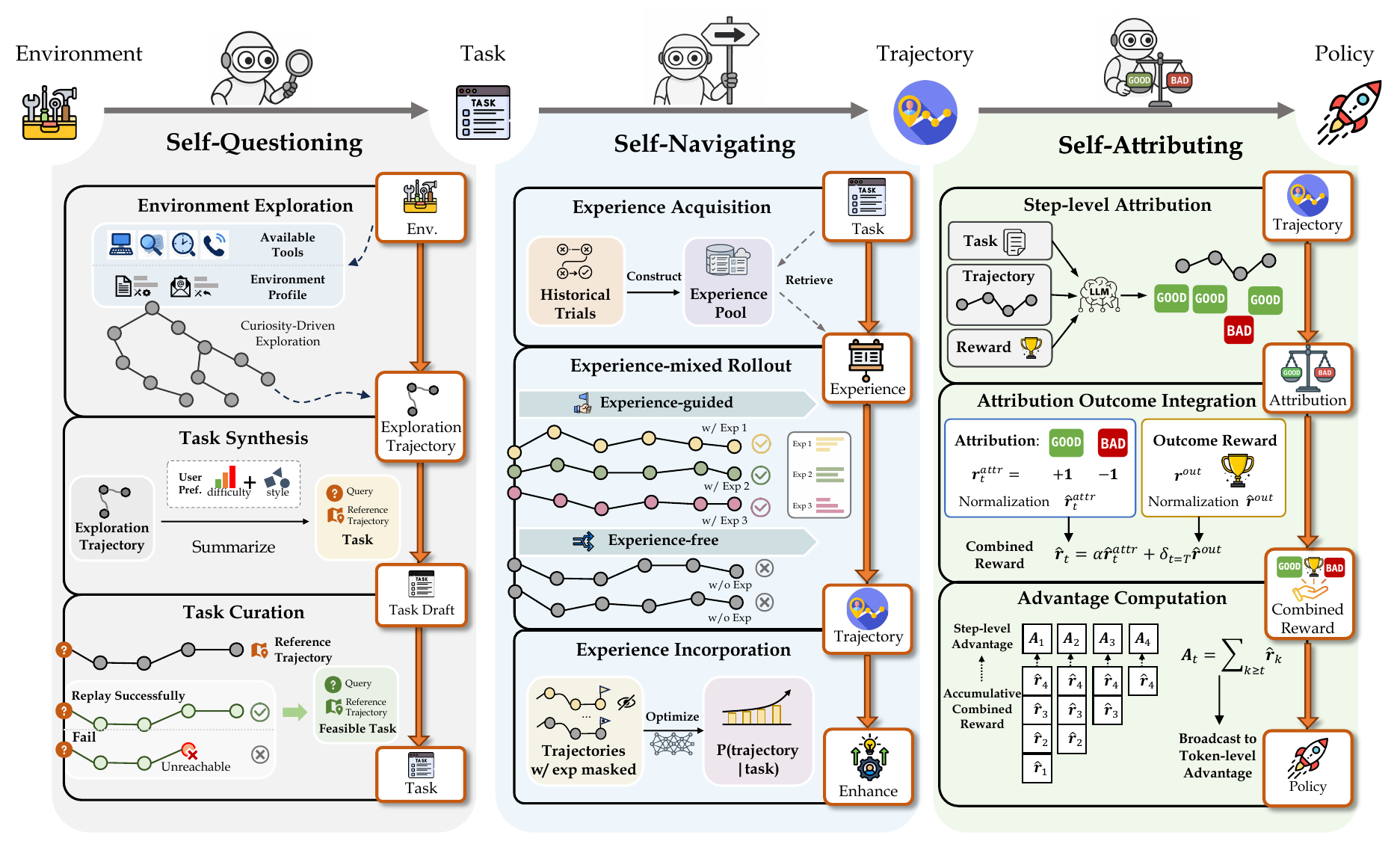}
  \caption{Overview of the \AgentEvolver framework. The self-evolving process is driven by three synergistic mechanisms: \textbf{\texttt{Self-questioning}} for autonomous task generation, \textbf{\texttt{Self-navigating}} for experience-guided exploration, and \textbf{\texttt{Self-attributing}} for fine-grained credit assignment.}
  \label{fig:agentevolver_overview}
\end{figure}

Specifically, \AgentEvolver aligns with the typical training flow—i) \emph{from environments to tasks}, ii) \emph{from tasks to trajectories}, and iii) \emph{from trajectories to agent policy}. 
As illustrated in Figure~\ref {fig:agentevolver_overview}, this self-evolution process comprises three synergistic mechanisms: \texttt{self-questioning}, \texttt{self-navigating}, and \texttt{self-attributing}:
\begin{itemize}
    \item \textbf{\texttt{Self-questioning}:} enables curiosity-driven exploration, allowing the LLM to autonomously generate tasks by probing the environment's state-action space and discovering functional boundaries. This mechanism reduces dependence on costly handcrafted datasets.
    \item \textbf{\texttt{Self-navigating}:} improves exploration efficiency by reusing and generalizing from past experiences. Through hybrid policy learning and trajectory guidance, the agent achieves more targeted and efficient task completion.
    \item \textbf{\texttt{Self-attributing}:} enhances sample efficiency by allocating fine-grained rewards along long trajectories. Instead of uniformly attributing outcomes as in typical GRPO methods, the LLM infers the respective contribution of intermediate states and actions, enabling more fine-grained and effective learning.
\end{itemize}
Together, these mechanisms constitute an integrated framework for agent self-evolution, systematically addressing the challenges of task scarcity, inefficient exploration, and low sample utilization. By shifting the training initiative from human-engineered pipelines to LLM-guided self-improvement, \AgentEvolver establishes a new paradigm that paves the way toward scalable, cost-effective, and continually improving intelligent systems.

Beyond the conceptual framework, we further introduce a practical infrastructure with a set of developer-oriented features. In particular, it ensures \emph{environmental compatibility} through standardized interfaces, enabling seamless interaction with diverse external environments. A unified \emph{Context Manager} then controls the agent’s multi-turn interaction logic under a consistent interface. It also integrates with reinforcement learning infrastructures such as \texttt{veRL} \citep{sheng2024hybridflow}, supporting efficient policy optimization and parameter updates. Moreover, the modular architecture facilitates secondary development: individual components can be optimized, extended, or redesigned to meet evolving research and application demands. In this way, \AgentEvolver serves both as a methodological contribution and as a practical foundation for building the next generation of self-evolving agent systems.

In the following sections, we first introduce the problem formulation of this work in Section~\ref{sec:formulation}.
We then present the three core mechanisms of the proposed framework:
Section~\ref{sec:self-questioning} presents the \texttt{self-questioning} module for curiosity-driven task generation; 
Section~\ref{sec:self-navigating} details the \texttt{self-navigating} strategy for efficient exploration; 
and Section~\ref{sec:self-attribution} explains the \texttt{self-attributing} mechanism for fine-grained credit assignment. 
Section~\ref{sec:infrastructure} then describes the supporting infrastructure. 
We evaluate the effectiveness of \AgentEvolver in Section~\ref{sec:experiments}, and finally conclude the report in Section~\ref{sec:conclusion} by summarizing the key findings and discussing future research directions.

\section{Problem Formulation}
\label{sec:formulation}

To establish a rigorous foundation for \textbf{agent self-evolution}, we formalize the learning problem as enabling an LLM-based agent to autonomously improve its policy through interaction with the environment---\textit{without relying on predefined task distributions or reward functions}. 
Our approach conceptually separates this problem into two components: (1) an interaction sandbox $\mathcal{E}$, which defines the environment dynamics (states, actions) but provides no objectives, and (2) an unknown target task distribution $p_{\text{target}}(g)$, representing the `golden' tasks the agent must ultimately master. The core challenge is thus for the agent to leverage open-ended exploration in $\mathcal{E}$ to effectively estimate $p_{\text{target}}(g)$, and in turn, autonomously generate its own proxy training objectives (tasks) and learning signals (rewards) to guide continual improvement. We now formally define these components.

\paragraph{Environment Formulation}
{
To distinguish our setting from standard RL environments, 
we define an \textit{interaction sandbox} that specifies the agent’s accessible state space, 
action space, and transition dynamics, but does not provide any predefined reward function or task objective. 
Formally, the interaction sandbox is represented as:
\begin{equation}
\label{eq:interaction_world}
\mathcal{E} = (\mathcal{S}, \mathcal{A}, \mathcal{P}),
\end{equation}
where $\mathcal{S}$ denotes the set of observable states, 
$\mathcal{A}$ the set of executable actions, 
and $\mathcal{P}(s' \mid s, a)$ the transition probability distribution governing the environment dynamics. 
Unlike a standard MDP environment ${E} = (\mathcal{S}, \mathcal{A}, \mathcal{P}, r, \gamma)$, 
the interaction sandbox $\mathcal{E}$ excludes the reward function $r$ as well as the discount factor $\gamma$, 
reflecting an open-ended, non-rewarding setting in which the agent must autonomously generate 
its own learning signals and objectives to drive continual policy improvement.}

\paragraph{Target Task Space and Oracle Reward}

The agent is ultimately evaluated on a target task distribution $p_{\text{target}}(g)$ over tasks $g \in \mathcal{G}$, where $\mathcal{G}$ denotes the task space. Each task $g$ is associated with a desired goal state $s_g \in \mathcal{S}$, representing a desired terminal state or outcome (e.g., a correct calculation, a retrieved fact, or a completed workflow). Intuitively, executing a task $g$ means steering the environment from an initial state toward its corresponding goal state $s_g$.

For each task $g$, a ground-truth reward function $R_g(s, a)$ measures the utility of taking action $a$ in state $s$ with respect to achieving the goal $s_g$. Starting from an initial state $s_0 \sim p_0$, the agent aims to learn a goal-conditioned policy $\pi_\theta(a \mid s, g)$ that maximizes the expected return across the target task distribution:
\begin{equation}
\label{eq:target_objective}
J_{\text{target}}(\theta) = 
\mathbb{E}_{g \sim p_{\text{target}},\, s_0 \sim p_0}
\left[ 
V^{\pi_\theta}(s_0, g)
\right],
\end{equation}
where the goal-conditioned value function follows the standard formulation \citep{schaul2015universal, liu2022goal}:
\begin{equation}
\label{eq:value_function}
V^{\pi_\theta}(s_0, g) 
= 
\mathbb{E} 
\left[ 
\sum_{t=0}^{\infty} \gamma^t R_g(s_t, a_t) 
\,\middle|\, 
s_0, g, \pi_\theta 
\right],
\end{equation}
which generalizes value estimation across both environmental states and goal-conditioned tasks.

\paragraph{Proxy Objective of Self-Evolution}
Since the target task distribution and reward functions are unknown a priori, the agent must synthesize its own training tasks and reward signals through formal generation mechanisms:

\begin{itemize}
    \item \textbf{Proxy Task Generation:} The agent explores the environment via a \texttt{self-questioning} mechanism (Section~\ref{sec:self-questioning}) to produce a candidate set of solvable tasks $\mathcal{G}_{\text{candidate}}$. This process can be formalized as a mapping from the environment to a task distribution:
    \begin{equation}
    \label{eq:task_mapping}
    F_{\text{task}}: \mathcal{E} \to \Delta(\mathcal{G}), \quad \text{where} \quad p_{\text{train}} = F_{\text{task}}(\mathcal{E})
    \end{equation}
    The resulting distribution $p_{\text{train}}(g)$ emphasizes tasks of appropriate difficulty and diversity, emulating the unknown $p_{\text{target}}(g)$.

    \item \textbf{Proxy Reward Design:} The agent infers a proxy reward function through \texttt{self-attributing} (Section~\ref{sec:self-attribution}), formalized as a mapping from environment and tasks to reward functions:
    \begin{equation}
    \label{eq:reward_mapping}
    F_{\text{reward}}: \mathcal{E} \times \mathcal{G} \to (\mathcal{S} \times \mathcal{A} \to \mathbb{R}), \quad \text{where} \quad \hat{R}_g = F_{\text{reward}}(\mathcal{E}, g)
    \end{equation}
    This enables finer-grained credit assignment than sparse environment rewards, allowing the agent to learn from richer feedback without human intervention.
\end{itemize}

The core objective of self-evolution is to formulate and optimize $F_{\text{task}}$ and $F_{\text{reward}}$ functions such that maximizing the proxy training objective:
\begin{equation}
\label{eq:train_objective}
J_{\text{train}}(\theta) = \mathbb{E}_{g \sim F_{\text{task}}(\mathcal{E})} \left[ \hat{V}^{\pi_\theta}(s_0, g) \right],
\end{equation}
with 
\begin{equation}
\label{eq:proxy_value}
\hat{V}^{\pi_\theta}(s_0, g) = \mathbb{E} \left[ \sum_{t=0}^{\infty} \gamma^t F_{\text{reward}}(\mathcal{E}, g)(s_t, a_t) \,\middle|\, s_0, g, \pi_\theta \right]
\end{equation}
leads to the improvement of the true target objective in Eq. \ref{eq:target_objective}. Moreover, it optimizes exploration through a self-navigating mechanism, guiding $\pi_\theta$ to generate high-learning-value trajectories and addressing efficiency bottlenecks in multi-turn environments.

In other words, this manuscript aims to formulate $F_{\text{task}}$ and $F_{\text{reward}}$ through systematic methodological development such that the policy $\pi_\theta$ optimized under the proxy tasks and rewards performs well on the true target task distribution. This formulation establishes a systematic framework where the agent's self-improvement is guided by carefully formulated task and reward generation mechanisms as well as efficient learning methods.

\section{Self-Questioning}
\label{sec:self-questioning}

\begin{figure}
  \centering
  \includegraphics[width=1.0\textwidth]{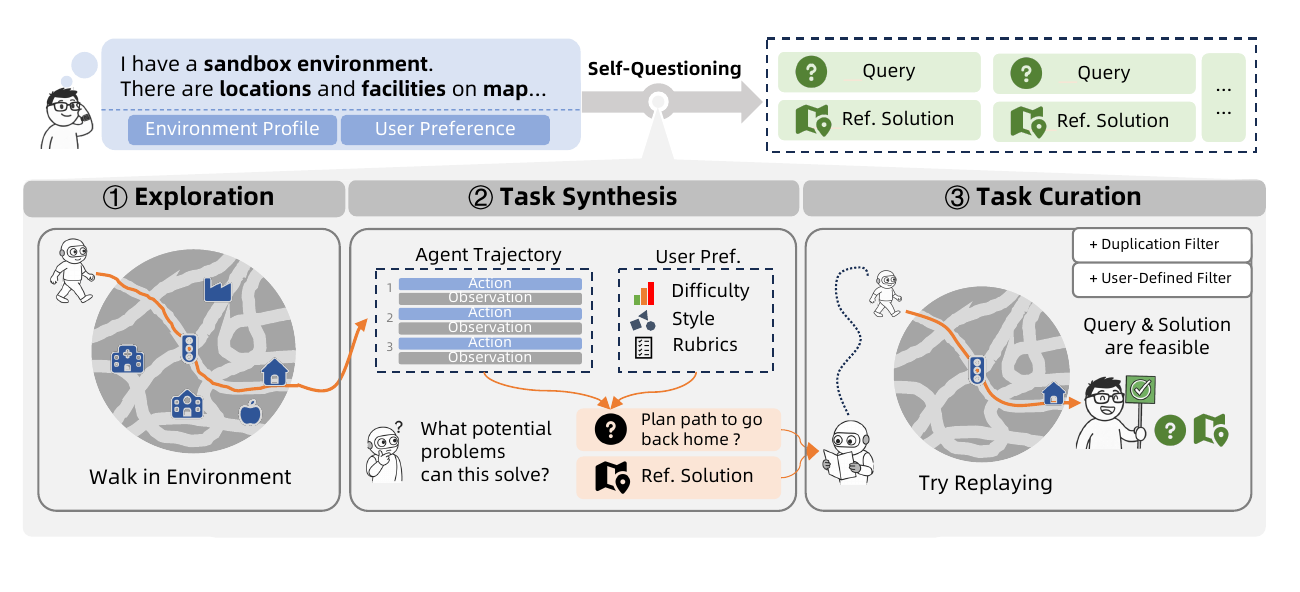}
  \caption{The pipeline of \texttt{self-questioning} module, including \textit{exploration}, \textit{task synthesis}, and \textit{task curation}. Initially, the module performs an exploratory phase across environments, which is directed by both environment profiles and user preferences (detailed in Section \ref{sec:exploration-with-profiles}). The generated trajectories are analyzed to formulate potential queries alongside their associated reference solutions. Next, we involve an agent that replays this path to verify the feasibility.}
  \label{fig:overview-cstb}
\end{figure}

In agent self-evolution, the adaptation process can be characterized as policy optimization over trajectories under proxy-generated tasks and rewards. Consequently, the quality of the self-synthesized tasks directly influences the effectiveness of policy improvement and its eventual transfer to the true target tasks.
However, in practical settings, acquiring training tasks of high quality remains challenging \citep{kang2025demystifying, villalobos2024position}. The difficulties mainly arise from two factors: i) Absence of training data in new environments: Most environments lack predefined data suitable for training, and manually constructing multi-step, complex tasks for agent learning is costly \citep{tang2025synthesizing}. ii) Low quality of training data: Available data frequently suffer from restricted diversity and skewed distributions, which constrain the agent's capacity to develop general and robust capabilities \citep{chen2024diversity}.

To address these challenges and enable agents to continually self-optimize across diverse environments, we propose a \texttt{self-questioning} mechanism
, which serves as the instantiation of the proxy task generation function $F_{\mathrm{task}}$ and synthetic rewards. As shown in Figure~\ref{fig:overview-cstb}, it implements a general exploration-generation pipeline to transform the sandbox $\mathcal E$ into exploration trajectories, and ultimately into user-preferred tasks, leveraging them to bootstrap self-evolution. The \texttt{self-questioning} module tries to answer four main problems in data generation:
\begin{enumerate}
    \item How to understand an unknown environment and find valuable states in it?  (Section \ref{sec:exploration-with-profiles})
    \item How to generate the tasks aligned with user preferences while maintaining sufficient diversity to cover heterogeneous demands? (Section \ref{sec:adaptive-task-synthesis})
    \item How to avoid hallucinations and guarantee the task executability? (Section \ref{sec:task-curation})
    \item How to provide proxy reward signals that evaluate agent performance at the trajectory level? (Section \ref{sec:self-questioning-llm-judge})
\end{enumerate}
By tackling these problems, the proposed module facilitates the co-evolution of tasks and agents, thereby progressively unlocking the agent's potential.

\subsection{Curiosity-Guided Exploration with Environment Profiles} \label{sec:exploration-with-profiles}

To effectively bootstrap synthetic tasks and enable agents to evolve corresponding abilities in a new environment, it is crucial to first develop a structured understanding of the unknown environment. To achieve that, we introduce curiosity-guided exploration with \textit{environment profiles}. The agent leverages these profiles to direct its curiosity-driven behaviors towards promising regions of the environment, explore diverse actions and states, and uncover the latent intentions in them.

\paragraph{Environment Profiles as Action Priors} 
Complex environments that agents are often involved in, e.g., browsers and editors, contain a wide variety of actions and states. However, users typically interact with only a subset that aligns with their specific tasks or preferences. To effectively acquire the relevant abilities to handle these problems, we capture such regularities through \textit{environment profiles}, which summarize entities, their attributes, and supported operations at the desired level of granularity.

\begin{center}
\begin{tcolorbox}[
  title=An example of \textit{Environment Profile},
  width=16cm,
  boxrule=0.45mm,
  fonttitle=\bfseries,
  enhanced,
  sharp corners,
  colback=white,
  colframe=black!70,
  arc=0mm,
  pad at break=2mm,
  outer arc=0mm
]
\texttt{%
\textbf{Entity: Map} \\
a game map with navigable roads and several points of interest, including a hospital, school, factory, home (or houses), and traffic lights.
\\
\textbf{Attributes:} \\
- road: a traversable pathway. \\
- architecture: a structure (e.g., hospital, school, home, or factory). \\
- traffic\_light: a signal requiring agents to wait before crossing. \\
\textbf{Operation:} \\
- move: move to an adjacent location.  \\
- wait\_and\_cross: wait for the signal to change and cross the intersection. \\
- enter: enter the nearest building.
}
\end{tcolorbox}
\captionof{figure}{An example of \emph{environment profile} of the sandbox in Figure \ref{fig:overview-cstb}. The elements in the map are represented by entities and attributes, while the operations enumerate possible conceptual actions for agents to build a fundamental understanding.} \label{fig:environment-profile}
\end{center}

During exploration, profiles serve as part of the initial state $s_0$ and help the LLM focus its curiosity and avoid aimless walking.

\paragraph{LLM-Driven Stochastic Exploration} We employ a high-temperature LLM as the primary source of curiosity. The elevated temperature encourages the agent to discover unconventional interaction patterns and hidden environment states, generating diverse and creative actions. At each step of exploration, the LLM observes the environment, thinks for the goals and options, and outputs an action to interact with the environment.

Formally, given the initial state $s_0$. At each step $t$, the agent samples an action from a profile-guided, high-temperature policy and then transitions according to environment dynamics:
\begin{equation}
a_t \sim \pi_\mathrm{explore}(\cdot\mid s_t,s_{t-1},s_{t-2},\dots,s_{0}),\qquad 
s_{t+1}\sim \mathcal{P}(\cdot\mid s_t,a_t),
\end{equation}
yielding a trajectory $\tau=(s_0,a_0,\ldots,s_T)$ with distribution $\rho(\tau)$.

During the exploration, we implement a two-phase exploration strategy to balance the efficiency with step allocation and myopic decision. During the initial $N_b$ steps, the agent conducts breadth-first exploration to establish a naive understanding of the environment's basic semantics, ensuring coverage of diverse regions of the action-state space. Subsequently, the agent transitions to depth-first exploration for more focused investigation of promising trajectories discovered earlier.
In this second phase, we further prevent premature convergence to a single behavior pattern by enforcing a myopic decision-making rule. Concretely, the agent only considers the most recent $N_d$ observations in $s_{t},s_{t-1},\dots,s_{t-N_d+1}$.

Through the aforementioned mechanism, we establish a transformation from the environment $\mathcal E$ to a distribution over trajectories. This process is guided by the LLM-driven stochastic policy $\pi_{\mathrm{explore}}$, which can be formalized as
\begin{equation}
    \Phi: \mathcal{E}\times\pi\times\mathcal S \to \mathcal{T}, \quad \text{where} \quad \rho = \Phi(\mathcal{E},\pi_\mathrm{explore},s_0),
\end{equation}
where $\mathcal{T}$ denotes the set of trajectories, $\rho(\tau)$ is explored with $\pi_{\mathrm{explore}}$ in $\mathcal E$, and $s_0$ concludes the environment profile.

\subsection{Adaptive Task Synthesis} \label{sec:adaptive-task-synthesis}

Having understood the environment through curiosity-guided exploration, in this section, the next step involves synthesizing training tasks that leverage the discovered environment capabilities. To achieve that, we propose adaptive task synthesis, which transforms the explored trajectories into tasks by combining the trajectory with user preferences to generate potential corresponding queries, and generating a reference solution based on the action-observation pairs.

\paragraph{Preference Guided Task Synthesis}

A key challenge in task synthesis is ensuring alignment with user preferences. To this end, we utilize \textit{user preferences} to provide structured guidance, constraining the task domain along two primary axes: task difficulty and task style.
Task difficulty is quantified by the number of entities, attributes, and operations involved in potential solutions. We adaptively control this by hierarchically scaling these factors. For instance, operations requiring coordination across multiple attributes are considered complex, while those involving few are simple. In addition to difficulty, our pipeline accepts user-defined rubrics to specify stylistic preferences, ensuring tasks reflect user demands.

This synthesis process initiates by processing raw exploration trajectories $\rho$. These trajectories are first distilled to extract salient interaction components: user inputs, LLM-generated actions, and their execution results. This distilled information is then augmented with the user preferences $u$ and provided to an LLM, which analyzes the context to synthesize a candidate query $g$ aligned with the desired constraints.

Formally, the transformation from experience to structured tasks through a preference-guided synthesis function is
\begin{equation}
    \Psi: \mathcal{T} \times \mathcal{U} \to \mathcal{G}, \quad \text{where} \quad g = \Psi\left(\rho, u\right) \sim p_\mathrm{train},
\end{equation}
where $\mathcal{U}$ denotes the space of user preferences $u$. By composing $\Phi$ and $\Psi$, we transform the environment $\mathcal E$ into its corresponding tasks $\mathcal G$ and implement $F_{\mathrm{task}}(\mathcal E) = \Psi(\Phi(\mathcal E),u)$, which yields the training task $g$ that satisfies distribution $p_{\mathrm{train}}$.

\paragraph{Reference Solution Extraction}

An essential part of our task synthesis is the extraction of reliable reference solutions. For problems posed in an unexplored environment, the ability of the agent would determine whether it could obtain a solution. In our \texttt{self-questioning} mechanism, however, problems are generated after exploration. This decoupling implies that an agent is presented with tasks whose solutions lie well beyond its current capability, and the correct answers are discoverable through the prior exploration phase. We leverage this property by designating these solutions as proxy ground-truth, i.e. reference solutions, which is shown in Figure~\ref{fig:overview-cstb}.
First, the explored trajectory is condensed into a simplified action-observation format. Then, this simplified trajectory and its corresponding task are provided to a prompted LLM to extract the complete solution process, ultimately yielding the reference solution. This solution encompasses all key steps required to solve the task.

By comparing the agent's predictions against the reference, we can robustly check correctness and provide targeted supervision.
In this way, we produce tasks of controlled difficulty and style, and ensure the availability of trustworthy solutions.

\subsection{Task Curation and Distributional Hybrid} \label{sec:task-curation}

Since $F_{\mathrm{task}}(\mathcal{E})$ produces a proxy task distribution $p_{\mathrm{train}}(g)$, the resulting samples inevitably vary in quality and difficulty. A systematic filtering mechanism is required to refine this distribution and ensure that $p_{\mathrm{train}}$ better emulates the unknown $p_{\mathrm{target}}$. To address the challenges, we propose a multi-layer filtering pipeline that evaluates the tasks across multiple dimensions during and after synthesis.

\paragraph{Real-time Filtering} During the task synthesis, the real-time filtering is applied to prevent the low-quality tasks. It operates on two primary criteria to avoid duplications. We first employ lexical overlap-based deduplication to eliminate redundant tasks. We compute token-level similarity scores between newly generated tasks and existing ones. Tasks exceeding a predefined threshold are immediately discarded in the early stage. The lexical filtering provides a computationally efficient first line of defense against redundancy. Next, semantic similarity-based check is introduced to further steer from duplicate tasks. Using embedding models, we maintain the representation of all existing tasks and guide the generation process to avoid similar trajectories.

\paragraph{Post-generation Filtering} After the initial synthesis and real-time filtering, we apply the post-filtering to ensure task quality and feasibility. This stage employs a more sophisticated evaluation. The post-filtering begins with the lexical deduplication, similar to the one in real-time filtering. Subsequently, we conduct a feasibility assessment to strictly verify task executability within the target environment. This evaluation leverages reference solutions obtained during the task synthesis phase and attempts to execute the solutions against the environment. By identifying real execution failures, potential mismatches between task requirements and environment, we effectively flag and remove synthetic tasks where the reference solution is likely hallucinatory.

\paragraph{Distributional Hybrid (Optional)}
In the default setting of interest where no external data are available, the agent relies entirely on proxy tasks, i.e.,
\begin{equation}
    p_{\mathrm{train}}(g) = F_{\mathrm{task}}(\mathcal{E})(g).
\end{equation}
Considering the superiority of mixed data in maintaining stable capabilities and expanding performance boundaries, if samples from the true target distribution $p_{\mathrm{target}}$ are available, we instead consider a hybrid distribution:
\begin{equation}
    p_{\mathrm{hybrid}}(g) = (1-\lambda) p_{\mathrm{target}}(g) + \lambda p_{\mathrm{task}}(g),
\end{equation}
with $\lambda \in (0,1]$ controlling the contribution of proxy tasks.

\subsection{Synthetic Task Reward with LLM Judge}\label{sec:self-questioning-llm-judge}

The synthesis of tasks serves as the initial step in our \texttt{self-questioning} module. Following this, we immediately face the second challenge: reliable task reward, which is necessary for guiding agent improvement. The difficulty is further acute in complex scenarios due to the lack of ground-truth rewards.
We address this challenge by developing a reward module and providing our general reference-based LLM judge for common automated assessment of task completion quality.

\paragraph{LLM as a Judge} The LLM Judge is a broadly validated solution for self-evaluation; however, designing one for long interaction trajectories remains a challenge. To accommodate developers, we offer an LLM-based judge as a fallback proxy for $F_{\mathrm{reward}}$, providing general-purpose assessment across diverse environments without specific optimizations but with agentic abilities. Our reward module ensures both generality and practicality for users to adopt their own environment rewards.

\paragraph{Basic Principle} To better assess a long-term trajectory, our LLM judge follows two principle to carefully assign rewards:

i) \textit{Relevance and Repetition Check} serves as the primary guard, identifying trajectories that contain steps unrelated to the target task, and all hallucinated or redundant steps. Agents that exhibit those behaviors without clear justification receive zero scores immediately. 

ii) \textit{Continuous Scoring} provides nuanced assessment beyond binary success metrics. Correct solutions receive scores in high range, while incorrect attempts fall within low range, with the specific score determined by factors such as step efficiency and intermediate errors. This graduated scoring captures partial progress and provides informative feedback for agent improvement.

\paragraph{Reference-based Correctness Check} Rather than relying solely on LLM knowledge to assess agents, our judge leverages reference solutions obtained in the exploration phase. It compares agent trajectories against reference solutions to verify coverage of essential steps. The evaluation checks whether all critical steps in the reference solutions appear in the agent's execution, either directly or through functionally equivalent alternatives.

This method allows us to maintain an objective standard for rewarding. Experimental validation of the effectiveness is presented in Section \ref{sec:experiment_self-questionin}.

\section{Self-Navigating}
\label{sec:self-navigating}

Efficient exploration in complex environments remains a central challenge for autonomous agents~\citep{lu2024intelligent, qiao2024agent}. Conventional RL approaches often rely on repetitive and random trial-and-error, resulting in redundant trajectories and slow convergence~\citep{burda2018exploration}. In contrast, human learning excels at extracting knowledge from past failures and systematically applying it to future endeavors. Motivated by this observation, we present a \textbf{\texttt{self-navigating}} mechanism that integrates experience reuse, trajectory guidance, and hybrid policy learning to improve \textbf{exploration efficiency}. \texttt{Self-navigating} enables agents to internalize and exploit prior experiences, thereby shifting exploration from unguided trial-and-error toward a more structured and scalable self-improvement process.

\begin{center}
\begin{tcolorbox}[
  title=An example of \emph{experience} for AppWorld,
  width=16cm,
  boxrule=0.45mm,
  fonttitle=\bfseries,
  enhanced,
  sharp corners,
  colback=white,
  colframe=black!70,
  arc=0mm,
  pad at break=2mm,
  outer arc=0mm
]
\texttt{%
\textbf{When to use:} \\ 
When attempting to use an API that hasn't been explicitly confirmed to exist or function as expected. \\
\\
\textbf{Content:} \\
Always verify the existence and behavior of an API through its specifications (using apis.api\_docs.show\_api\_doc) before executing calls, especially for critical actions like deletions or modifications.
}
\end{tcolorbox}
\label{fig:experience notion}
\captionof{figure}{An example of \emph{experience} for AppWorld, consisting of two components: \emph{When to use} and \emph{Content}.}
\end{center}

\paragraph{Notion of Experience}

In the context of agent self-evolution, an \emph{experience} encapsulates distilled insights derived from past trajectories, encompassing both successful and unsuccessful cases. In our setting, we formalize an \emph{experience} as a unit of structured knowledge articulated in natural language, designed to guide agents toward appropriate decisions through In-Context Learning (ICL). Each \emph{experience} comprises two components: \emph{When to use} and \emph{Content}. The \emph{When to use} component specifies the conditions under which the experience is applicable and serves as a retrieval trigger, with queries semantically matching to its vectorized embedding. The \emph{Content} component provides detailed instructions, precautions, or recommended strategies for effective action within that context. Representing experiences entirely in natural language preserves interpretability while supporting scalable, embedding-based retrieval.

\subsection{Experience Acquisition}\label{sec:exp_acq}
To enable effective experience reuse, we introduce an \emph{experience acquisition} stage that spans both offline and online phases. In the offline phase, the \emph{Pool Construction} process builds a diverse and well-structured repository of high-quality experiences. During the online phase (i.e., rollout or inference), the \emph{Experience Retrieval} process dynamically draws task-specific knowledge from this repository to provide contextually relevant guidance. Together, these two processes ensure that prior trajectories can be utilized effectively by constructing experiences while preserving adaptability across diverse tasks~\footnote{More implementation details can be found at: \url{https://github.com/agentscope-ai/ReMe}.}.

\begin{figure}
  \centering
  \includegraphics[width=1.0\textwidth]{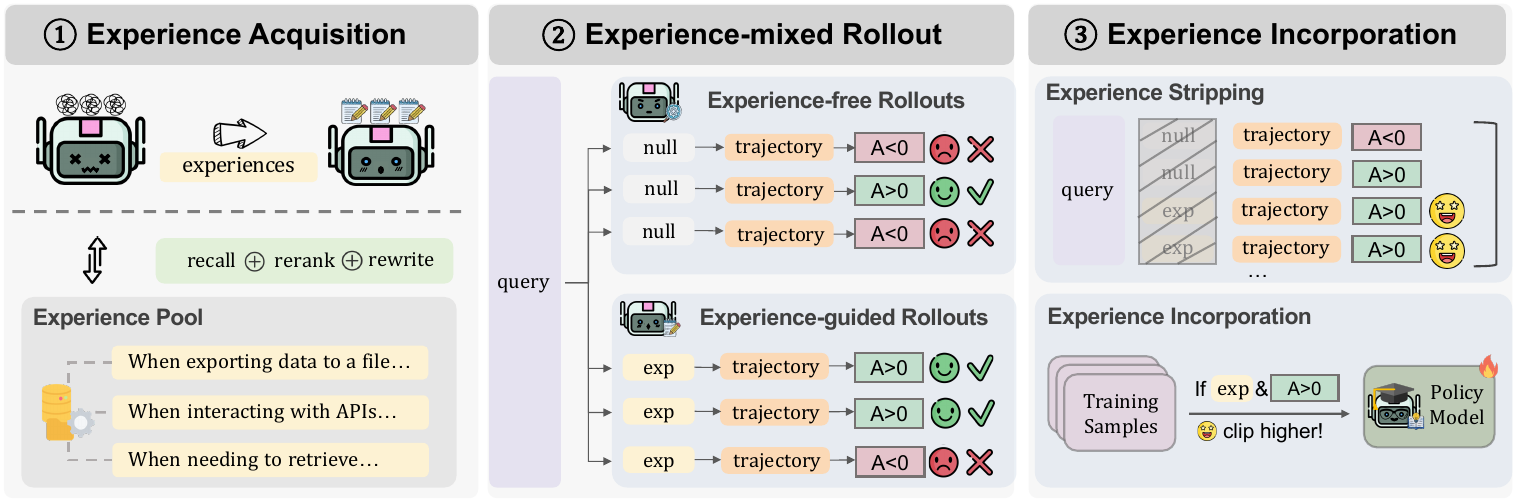}
  \caption{The pipeline of \texttt{self-navigating} mechanism. It comprises three key stages: (i) \emph{Experience Acquisition}, where the agent constructs and retrieves structured natural-language experiences distilled from prior trajectories; (ii) \emph{Experience-mixed Rollout}, which interleaves unguided and experience-guided trajectories to balance exploration and exploitation; and (iii) \emph{Experience Incorporation}, which integrates retrieved experiences into the policy optimization process.}
  \label{fig:navigating}
\end{figure}

\paragraph{Pool Construction}

To construct a cold-start experience pool $\mathcal{P}_{\text{exp}}$ for environment $\mathcal{E}$ using the proxy task distribution $p_{\text{train}}$ (derived from the \texttt{self-questioning} module in Section~\ref{sec:self-questioning}), we proceed as follows. For each task $g$ sampled from $p_{\text{train}}$, the initial policy $\pi_{\theta_{\text{init}}}$ performs $N_{\text{pc}}$ independent rollouts in $\mathcal{E}$, producing a set of trajectories. Each trajectory set then undergoes through a processing pipeline $\Omega_{\text{process}}$, which consists of the following key stages:
(1) \emph{trajectory preprocessing}, where collected trajectories are validated and classified into success and failure groups according to their scores;
(2) \emph{experience extraction}, where task memories are distilled from successful, failed, or comparative trajectory pairs, capturing essential behavioral insights;
(3) \emph{experience validation}, which employs an LLM-based evaluator to assess the quality and reliability of each extracted experience; and
(4) \emph{vector store updating}, where validated task experiences are indexed and integrated into the long-term experience storage for subsequent retrieval and reasoning.
The final experience pool $\mathcal{P}{\text{exp}}$ is obtained by aggregating the processed trajectory sets across all sampled tasks, formally defined as:
\begin{equation}
\mathcal{P}_{\text{exp}} = \bigcup_{g \sim p_{\text{train}}} \Omega_{\text{process}} \left( \left\{ \tau_i^{(g)} \right\}_{i=1}^{N_{\text{pc}}} \right), \quad \text{where} \quad \tau_i^{(g)} \sim \pi_{\theta_{\text{init}}}(\cdot \mid \mathcal{E}, g).
\end{equation}

\paragraph{Experience Retrieval}
Given a task $g$, we extract its associated query $q$ and encode it into an embedding vector $\mathbf{h}_q$. Each experience $e \in \mathcal{P}_{\text{exp}}$ is represented similarly by an embedding $\mathbf{h}_e$. The similarity between $\mathbf{h}_q$ and all $\mathbf{h}_e$ is computed to retrieve the most relevant experiences. The top-$k$ most similar experiences are then refined through a module $\Omega_{\text{refine}}$, which performs (1) re-ranking to ensure contextual relevance and task alignment, and (2) re-writing to enhance the generality and adaptability of the retrieved experiences.

The overall retrieval process is formalized as:
\begin{equation}
    EXP(g) = \Omega_{\text{refine}} \Big( 
        \operatorname{TopK}_k \big( 
            \{ \operatorname{sim}(\mathbf{h}_q, \mathbf{h}_e) \mid e \in \mathcal{P}_{\text{exp}} \} 
        \big) 
    \Big),
\end{equation}
where $\operatorname{sim}(\cdot, \cdot)$ denotes the similarity measure computed by the cosine similarity function, $\operatorname{TopK}_k$ selects the $k$ experiences with the highest similarity scores, and $\Omega_{\text{refine}}(\cdot)$ performs re-ranking and re-writing to yield the final retrieved set $EXP(g)$.

\subsection{Experience-mixed Rollout}
While prior experiences offer valuable guidance, excessive reliance on them may hinder exploration. To balance this trade-off, we introduce an \emph{experience-mixed rollout} stage that interleaves unguided and experience-guided trajectories. By integrating these two types of rollouts, our approach allows agents to efficiently leverage prior knowledge while maintaining flexibility and adaptability in novel situations.

\paragraph{Trajectory Generation}
Given the task $g$, the policy model $\pi_{\theta_{\text{old}}}$ interacts with the environment to generate a set of trajectories $\mathcal{T}=\mathcal{T}^{(v)} \cup \mathcal{T}^{(e)}$.
Specifically, $\mathcal{T}^{(v)}$ denotes \textbf{vanilla rollouts}, which are generated solely by the policy model without external guidance; whereas $\mathcal{T}^{(e)}$ corresponds to \textbf{experience-guided rollouts}, wherein retrieved experience is incorporated to steer the generation process:
\begin{equation}
    \mathcal{T}^{(v)}=\left\{\tau_i^{(v)}\right\}_{i=1}^{N_v}, \quad \mathcal{T}^{(e)}=\left\{\tau_j^{(e)}\right\}_{j=1}^{N_e}, \quad N=N_v+N_e.
\end{equation}

The balance between the two types of rollouts is controlled by $\eta$, with $N_e=\lfloor \eta \cdot N \rfloor$. The higher $\eta$ values favor prior experience, while the lower values promote unconstrained exploration. For experience-guided rollouts, a relevant experience snippet $exp_g\in EXP(g)$ is first retrieved (see Section~\ref{sec:exp_acq}) and injected into the initial prompt following a predefined template, e.g.,  
\texttt{\colorbox{gray!10}{\{system\_prompt\}<EXP>\{$exp_g$\}</EXP>\{query\}}}.

\paragraph{Advantage Computation}
To ensure a consistent treatment of both vanilla and experience-guided trajectories, we compute the standardized advantage for each trajectory $\tau \in \mathcal{T}$ across the entire set:
\begin{equation}
\hat{A}(\tau)=\frac{R(\tau)-\mu_R}{\sigma_R+\varepsilon_{\text {norm }}},
\end{equation}
where $R(\tau)$ denotes the reward for trajectory $\tau$, 
$\mu_R$ and $\sigma_R$ represent the mean and standard deviation of rewards across all trajectories,
and $\varepsilon_{\text {norm }}$ is a small constant introduced for numerical stability.

This design allows the model to explicitly leverage prior textual experience through in-context learning, steering generation toward reliable and sample-efficient reasoning paths while maintaining flexibility with unconstrained vanilla rollouts. By integrating vanilla and experience-guided trajectories, the proposed experience-mixed rollout strategy strikes a balance between \textbf{exploration and exploitation}, offering a principled foundation for subsequent policy optimization.

\subsection{Experience Incorporation}
To seamlessly integrate the benefits of retrieved experiences into RL training, we propose an \emph{experience incorporation} mechanism that operates during the optimization stage. This mechanism comprises two complementary components: \emph{experience stripping} and \emph{selective boosting}. The former prevents spurious dependence on explicit textual cues, while the latter amplifies the influence of pertinent constructive experiences. In combination, these procedures ensure that experience effectively guides policy learning while mitigating the peril of over-reliance on external prompts.

\paragraph{Experience Stripping}

While experience-guided rollouts leverage extracted experience by injecting it into the input prompt, retaining these tokens during optimization risks undesirable overfitting, as the model may memorize external content rather than assimilating its underlying reasoning signals. To address this, we adopt an \emph{experience stripping} strategy: prior to policy gradient computation, retrieved experience tokens are removed from the training samples. This ensures that optimization is driven exclusively by the generated trajectory, thus attributing reward signals to the model’s intrinsic reasoning processes rather than auxiliary contextual cues. Consequently, the rollout phase continues to benefit from experience, whereas the training phase fosters genuine policy improvement.

\begin{center}
\begin{tcolorbox}[
  title=An illustration of \emph{Experience Stripping},
  width=16cm,
  boxrule=0.45mm,
  fonttitle=\bfseries,
  enhanced,
  sharp corners,
  colback=white,
  colframe=black!70,
  arc=0mm,
  pad at break=2mm,
  outer arc=0mm
]
\texttt{%
\textbf{Before:} \\ 
\{system\_prompt\}\colorbox{gray!15}{<EXP>\{experience\}</EXP>}\{query\}\{trajectory\} \\
\\
\textbf{After:} \\
\{system\_prompt\}\{query\}\{trajectory\}
}
\end{tcolorbox}
\captionof{figure}{Illustration of \emph{experience stripping}.}
\label{fig:experience stripping}
\end{center}

\paragraph{Selective Boosting}
After obtaining experience-guided training samples with the experience tokens removed, it is important to note that the original rollout trajectories were conditioned on both the query and the experience, whereas the current optimization samples are conditioned solely on the query. This inference and learning sample \textbf{mismatch} causes a typical off-policy learning problem where the probability distribution of the optimized samples is substantially narrower than that of the rollout samples~\citep{yan2025learning}, leading the importance ratio $r^{(e)}$ to grow uncontrollably.

Although standard PPO/GRPO clipping can partially mitigate this instability by constraining the importance ratio, excessive clipping suppresses the gradient contributions of experience-guided updates and thus limits learning efficiency. To overcome this limitation, we propose a \emph{selective boosting} strategy. As shown in Eq.~\eqref{eq:navigating_loss}, for samples with strictly positive advantages ($\hat{A}^{(e)} > 0$), the upper clipping threshold $\hat{\epsilon}_{\text{high}}$ is selectively boosted, allowing larger importance ratios to pass through without attenuation. This preserves the optimization signals from favorable experience-conditioned trajectories, enabling the policy model to learn more effectively from experience.

\begin{equation}\label{eq:navigating_loss}
\begin{aligned}
\mathcal{L}_{\text{navigating}}(\theta) = 
-\frac{1}{N} \Biggl[ &
\sum_{i=1}^{N_v} \min \Big( r_i^{(v)} \hat{A}_i^{(v)}, 
\operatorname{clip}\big( r_i^{(v)}, 1-\epsilon_{\text{low}}, 1+\epsilon_{\text{high}} \big) \hat{A}_i^{(v)} \Big) \\
+ & \sum_{j=1}^{N_e} \min \Big( r_j^{(e)} \hat{A}_j^{(e)}, 
\operatorname{clip}\big( r_j^{(e)}, 1-\epsilon_{\text{low}}, 1+\epsilon_j \big) \hat{A}_j^{(e)} \Big) 
\Biggr] 
+ \beta \operatorname{KL}\left( \pi_\theta \| \pi_{\theta_{\text{old}}} \right),
\end{aligned}
\end{equation}
\vspace{-0.2cm}
\begin{align*}
\text{where:} \quad 
& r_i^{(v)} = \frac{\pi_\theta(\tau_i^{(v)})}{\pi_{\theta_{\text{old}}}(\tau_i^{(v)})}, \quad 
r_j^{(e)} = \frac{\pi_\theta(\tau_j^{(e)})}{\pi_{\theta_{\text{old}}}(\tau_j^{(e)})}, \quad
\epsilon_j = 
\begin{cases} 
\hat{\epsilon}_{\text{high}} & \text{if } \hat{A}_j^{(e)} > 0 \\
\epsilon_{\text{high}} & \text{otherwise}
\end{cases} \quad.
\end{align*}

\section{Self-Attributing}
\label{sec:self-attribution}

In this section, we introduce the final stage of \texttt{AgentEvolver}, \texttt{self-attributing} mechanism.
Existing methods such as GRPO~\citep{deepseek_math} rely on sparse, trajectory-level rewards that equally credit all actions, failing to distinguish between critical decisions and inconsequential steps. This  credit assignment leads to inefficient sample utilization and limits the policy's ability to learn from fine-grained feedback.

To address this challenge, \texttt{self-attributing} transforms multi-step trajectories into efficient learning signals by leveraging an LLM's reasoning capabilities to assess the contribution of each action. It serves as the concrete implementation of our proxy reward function, $F_\mathrm{reward}$ in Eq.~\ref{eq:reward_mapping} (Section~\ref{sec:formulation}). We reformulate credit assignment by shifting from temporal signal propagation to \emph{contribution-based attribution}. It leverages a large language model's reasoning capabilities to retrospectively determine whether each action contributed positively or negatively to the final outcome. 

We structure the learning signal around two distinct dimensions: \textbf{process quality} (the contribution of an action step) and \textbf{outcome effectiveness} (the final trajectory success). To preserve the distinct contribution of each signal and prevent cross-signal interference, we apply standardization to each dimension separately before fusing them into a composite reward. This signal is then mapped to token-level advantages for GRPO optimization. By enabling precise, step-wise credit assignment, this design significantly enhances sample efficiency, supporting \texttt{AgentEvolver}'s objective of scalable and continual agent improvement. The entire process is illustrated in the Self-Attributing panel of Figure~\ref{fig:agentevolver_overview}.

\begin{figure}[htbp] 
\centering

\begin{tcolorbox}[
  title=System Prompt for Step-wise Attribution,
  width=\linewidth, 
  boxrule=0.45mm,
  fonttitle=\bfseries,
  enhanced,
  sharp corners,
  colback=white,
  colframe=black!70,
]
\setlist[itemize]{itemsep=0pt, topsep=2pt, parsep=0pt, partopsep=0pt, leftmargin=*}

You are an expert process reward evaluator specializing in attributional analysis.

\vspace{2mm}
\noindent \textbf{INPUTS:} A user's \texttt{TASK}, a numbered \texttt{SOLUTION TRAJECTORY} of steps, and a final \texttt{OVERALL PERFORMANCE SCORE}.

\vspace{1mm}

\noindent \textbf{YOUR TASK:} Attribute the contribution of each step in the trajectory to the final score based on the following rules.

\vspace{1mm}

\noindent \textbf{EVALUATION RULES:}
\begin{itemize} 
    \item \textbf{If Score is POSITIVE $>0$:}
        \begin{itemize} 
            \item \textbf{GOOD}: The step contributed positively to the solution.
            \item \textbf{BAD}: The step was irrelevant, neutral, or detrimental.
        \end{itemize}
    \item \textbf{If Score is NEGATIVE $\le 0$:}
        \begin{itemize}
            \item \textbf{GOOD}: The step \textbf{only if} it actively corrected or mitigated an error.
            \item \textbf{BAD}: The step introduced, propagated, or failed to fix an error.
        \end{itemize}
\end{itemize}

\vspace{1mm}
\noindent \textbf{FOCUS:} Evaluate strictly on the \textbf{technical impact} of each step's action. Ignore superficial elements like politeness. Reply in the required format only.
\end{tcolorbox}

\caption{System Prompt Structure for step-wise attribution.} 
\label{fig:attr_prompt_system}
\end{figure} 

\subsection{Step-wise Attribution via LLM-based Reasoning} \label{sec:llm_attribution}
The first step of the \texttt{self-attributing} pipeline is to generate the \textbf{process quality} signal (Figure~\ref{fig:agentevolver_overview}, Self-Attributing panel, top). This is accomplished by using an LLM's reasoning capabilities to retrospectively analyze completed trajectories. The goal is to produce a step-wise evaluation of each action's contribution to the final outcome, serving as a measure of its logical correctness within the task context.

To ensure contextual consistency and computational efficiency, we evaluate each trajectory holistically in a single pass. 
Specifically, we format the entire context—including the initial task, all intermediate steps, and the final outcome—into a single prompt (see Figures.~\ref{fig:attr_prompt_system} and~\ref{fig:attr_prompt_user}) for LLM-based evaluation. 
It allows the LLM to capture complex inter-step dependencies, ultimately outputting a binary label for each step: \texttt{GOOD} for beneficial actions and \texttt{BAD} for those deemed irrelevant or counterproductive.

This method of generating the \textbf{process quality} signal offers several advantages over conventional, handcrafted Process Reward Models (PRMs). First, by outputting binary labels (\texttt{GOOD}/\texttt{BAD}), it directly operationalizes the concept of attribution without requiring complex, task-specific scoring schemes, which are often difficult to design and transfer across tasks. Second, it leverages the LLM's high-level reasoning capabilities to make contextual judgments, replacing rigid heuristics with flexible, situation-aware analysis. Most importantly, the resulting attribution labels constitute a semantically-grounded measure of process correctness. They provide directional feedback (i.e., positive or negative contribution) rather than precise reward magnitudes. This qualitative signal is exactly what is needed to construct one of the two independent channels—process quality—before its combination with outcome effectiveness in the subsequent step (Section~\ref{sec:attribution_integration}).
This qualitative signal constitutes the \textbf{process quality} channel, one of two independent inputs to our final reward. Its integration with the 'outcome effectiveness' channel is detailed in Section~\ref{sec:attribution_integration}.

\begin{figure}[htbp] 
\centering 
\begin{tcolorbox}[
  title=User Prompt for Step-wise Attribution,
  width=\linewidth, 
  boxrule=0.45mm,
  fonttitle=\bfseries,
  enhanced,
  sharp corners,
  colback=white,
  colframe=black!70,
]
\textbf{TASK DESCRIPTION} \\
\texttt{<Original task query>}

\vspace{2mm}
\textbf{SOLUTION TRAJECTORY (total N steps)}
\begin{verbatim}
>>> EVAL-STEP 0 <<<
<ACTION> ... <END>
<OBSERVATION> ... <END>
\end{verbatim}
\textit{[…continue for all steps…]}

\vspace{2mm}
\textbf{OVERALL PERFORMANCE SCORE} \texttt{<value>}

\vspace{2mm}
\textbf{REQUIRED OUTPUT FORMAT:}
\begin{description}[style=unboxed, leftmargin=0pt, itemsep=1pt]
    \item[Step 0 Analysis:] \texttt{<your reasoning>}
    \item[Step 0 Judgment:] GOOD/BAD
    
    \textit{[…continue for all steps…]}
\end{description}
\end{tcolorbox}

\caption{User prompt Structure for step-wise attribution, which provides the concrete data for evaluation.} 
\label{fig:attr_prompt_user}
\end{figure}

\subsection{Attribution-Based Reward Construction}
\label{sec:attribution_reward}
Building upon the step-wise qualitative labels (GOOD/BAD) labels generated in Section~\ref{sec:llm_attribution}, this section details the procedure for constructing a quantitative, step-wise attribution reward, $r_t^{\text{attr}}$. 
This signal is designed to function as a dense, process-oriented feedback mechanism that complements the sparse, outcome-based reward. The goal is to convert the LLM's discrete assessment into a continuous signal suitable for policy optimization.

\paragraph{Quantification of Attribution Labels.}

To convert the categorical \texttt{GOOD}/\texttt{BAD} labels into a quantitative signal, we assign a numerical attribution reward, $r_t^{\text{attr}}$, to each step: $+1$ for \texttt{GOOD} and $-1$ for \texttt{BAD}. This binary assignment provides a clear directional signal of each step's localized contribution, without introducing arbitrary scalar magnitudes that could bias gradient computation. Unlike continuous scoring schemes that require careful calibration across diverse tasks and environments, this formulation remains interpretable and stable. These step-wise attribution scores will subsequently be normalized and integrated with outcome-based rewards (Section~\ref{sec:attribution_integration}), enabling the policy to optimize for both procedural correctness and task success.

\paragraph{Normalization of the Attribution Signal.} 
To ensure numerical stability and standardize the signal's scale before its integration with the outcome-based reward, we apply standardization to the raw attribution rewards, $r^{\text{attr}}_t$. A critical design choice is the population over which the normalization statistics---mean ($\mu^{\text{attr}}$) and standard deviation ($\sigma^{\text{attr}}$)---are computed. Since trajectories can vary significantly in length, a naive step-level calculation could allow longer, potentially meandering trajectories to disproportionately influence the statistics.

To address this and ensure that each trajectory is treated as an equally important trial, we adopt \textbf{trajectory-level standardization}. In this approach, we first compute the \emph{average} attribution reward for each trajectory. The mean ($\mu^{\text{attr}}$) and standard deviation ($\sigma^{\text{attr}}$) are then calculated over this population of trajectory-average rewards. This method gives each trajectory equal weight in the normalization process, regardless of its length, ensuring that the statistics reflect the distribution of overall trajectory quality rather than being skewed by the volume of individual steps.

The standardized attribution reward, $\hat{r}^{\text{attr}}_t$, is calculated for each step using these trajectory-level statistics:
\begin{equation}
    \hat{r}^{\text{attr}}_t = \frac{r^{\text{attr}}_t - \mu^{\text{attr}}}{\sigma^{\text{attr}} + \epsilon},
\end{equation}
where $\epsilon$ (e.g., $10^{-8}$) is a small constant to prevent division by zero.

Thus, $\hat{r}^{\text{attr}}_t$ serves as a dense, normalized reward signal grounded in the LLM's attribution of each action's correctness.

\subsection{Composite Reward Construction}
\label{sec:attribution_integration}

To form a complete learning signal, we integrate the attribution-based reward $\hat{r}^{\text{attr}}_t$ (from Section~\ref{sec:attribution_reward}) with an outcome-based reward. This dual-channel approach, termed Attribution Outcome Integration (Figure~\ref{fig:agentevolver_overview}, right panel, middle), allows the agent to optimize for both procedural correctness and final task success. The key to this fusion is the seperated normalization of each channel, which preserves their distinct contributions. This section details the formulation of this composite reward.

\paragraph{Outcome Reward Formulation and Normalization.}
To complement the dense, process-oriented attribution signal, we utilize the terminal reward in Section~\ref{sec:self-questioning-llm-judge} as $R^{\text{out}}$, that reflects the final outcome of a trajectory. 

To maintain statistical independence from the attribution channel, this trajectory-level outcome reward is normalized separately across the training group. The standardized outcome reward, $\hat{r}^{\text{out}}$, is computed as:
\begin{equation}
    \hat{r}^{\text{out}} = \frac{R^{\text{out}} - \mu^{\text{out}}}{\sigma^{\text{out}} + \epsilon},
\end{equation}
where $\mu^{\text{out}}$ and $\sigma^{\text{out}}$ are the mean and standard deviation of the raw outcome rewards $R^{\text{out}}$ across all trajectories in the group.
This independent normalization is crucial to prevent the scale of one signal from inadvertently dominating the other during fusion.

\paragraph{Reward Fusion.}
The final composite reward for each step $t$, denoted as $\hat{r}_t$, is then formulated by combining the dense, process-oriented attribution signal with the sparse, outcome-based signal. The outcome reward, $\hat{r}^{\text{out}}$, is applied only at the terminal step of the trajectory to correctly attribute the final success or failure. The formula is a weighted sum:
\begin{equation}
    \hat{r}_t = \alpha \cdot \hat{r}^{\text{attr}}_t + \mathbf{1}_{t=T} \cdot \hat{r}^{\text{out}}.
\end{equation}
Here, $\mathbf{1}_{t=T}$ is an indicator function that equals 1 if $t=T$ and 0 otherwise. The hyperparameter $\alpha \ge 0$ controls the relative contribution of the attribution signal.

\paragraph{Advantage Computation.}
Next, we compute the step-level advantage, $A_t$, from the composite rewards, $\hat{r}_k$. For this, we adopt a simplified yet effective advantage estimation method, following the precedent set by works like DeepSeek-Math~\citep{deepseek_math}. It defines the advantage directly as the undiscounted cumulative future reward. This formulation is a special case of the standard discounted return where the discount factor $\gamma$ is set to 1, thereby giving equal weight to all subsequent steps in a trajectory. The advantage is thus calculated as:
\begin{equation}
    A_t = \sum_{k=t}^{T} \hat{r}_k.
\end{equation}
This calculation produces a single advantage value for each step that cohesively integrates both the process correctness derived from LLM-based attribution and the empirical task success.

\paragraph{The Role of the Weighting Hyperparameter $\alpha$.}
The hyperparameter $\alpha$ provides a direct mechanism to balance the influence of process correctness against outcome maximization. A higher value for $\alpha$ biases the agent towards learning policies that align with the LLM's attribution assessments, fostering robust and generalizable reasoning procedures. Conversely, a lower value prioritizes achieving the final task goal.

This adjustable balance is particularly useful for implementing curriculum learning. For instance, training can commence with a higher $\alpha$ to establish a strong procedural foundation, and then gradually decrease $\alpha$ to fine-tune the policy for optimal task-specific performance.

\subsection{Policy Optimization}
\label{sec:grpo_integration}
The step-level advantages derived in Section~\ref{sec:attribution_integration} must be mapped to token-level signals for policy gradient optimization. This section details the mapping procedure and the resulting optimization objective (Figure~\ref{fig:agentevolver_overview}, right panel, bottom).

\paragraph{Step-to-Token Mapping.}
Policy gradient methods for language models require token-level advantage estimates. For each step $t$ corresponding to an action, we broadcast the step-level advantage $A_t$ to all response tokens generated during that step, yielding token-level advantages $A^{\text{tok}}_j$. Formally, if token $j$ belongs to step $t$ as determined by the step identification mechanism, then $A^{\text{tok}}_j = A_t$.

\paragraph{Integration with Experience-Guided Training.}
These token-level advantages $A^{\text{tok}}_j$ serve as the advantage estimates for both vanilla and experience-guided rollouts in the complete optimization objective (Equation~\ref{eq:navigating_loss}). The composite advantage term incorporates both attribution-based process evaluation and outcome-based task success, enabling the policy to optimize for both intermediate step quality and final trajectory outcomes. This unified treatment ensures consistent credit assignment across different trajectory types while maintaining the selective boosting mechanism for experience-guided samples described in Section~\ref{sec:self-navigating}.

\section{Framework and Infrastructure}
\label{sec:infrastructure}

\begin{figure}
  \centering
  \includegraphics[width=1.0\textwidth]{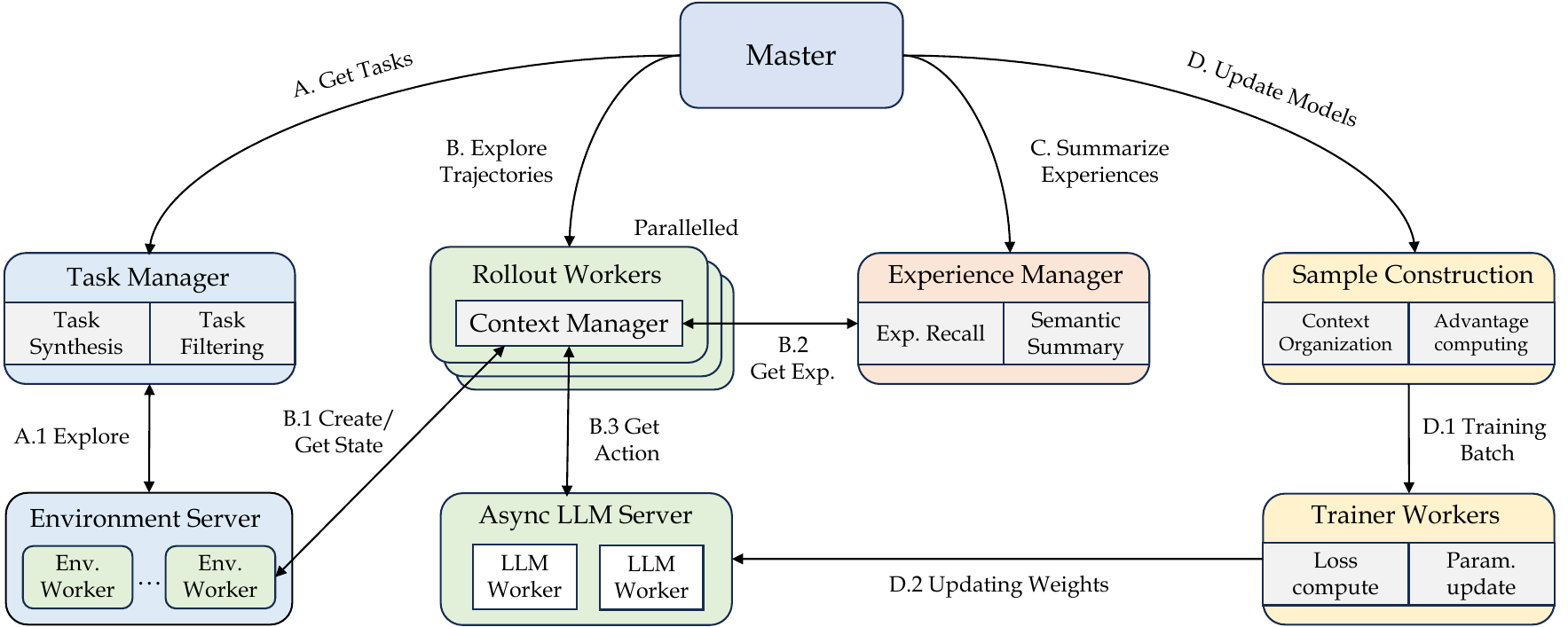}
  \caption{Framework of the AgentEvolver System.}
  \label{fig:infra_framework}
\end{figure}

\subsection{Training Framework}
\label{sec:framework}

This section introduces the training framework that turns the three mechanisms---\texttt{self-questioning}, \texttt{self-navigating}, and \texttt{self-attributing}---into an end-to-end, service-oriented system. The design emphasizes (i) a clear dataflow from environments to model updates, (ii) a hierarchical rollout stack tailored to agentic, multi-turn interaction, and (iii) algorithmic extensibility via modular, decoupled components.

\paragraph{Overall Architecture and Data Flow.}
As illustrated in Figure~\ref{fig:infra_framework}, a \emph{Master Orchestrator} drives a four-stage loop:
(A) \textbf{Task Synthesis}, (B) \textbf{Trajectory Rollout}, (C) \textbf{Experience Summarization}, and (D) \textbf{Sample Construction \& Model Optimization}.
First, \emph{Task Synthesis} is handled by a \emph{Task Manager}, which performs \textbf{self-questioning} by interacting with the environment to explore and automatically generate candidate training tasks.
\emph{Trajectory Rollout} then launches parallel executions of these tasks, where multiple rollout workers conduct multi-turn interactions between the model and the environment to generate diverse trajectories.
Next, \emph{Experience Summarization} is performed by the \textbf{Experience Manager}, which condenses historical trajectories into compact \emph{experience texts} (e.g., skills, tactics, or failure notes).
These summaries are indexed and retrieved during future rollouts to provide in-context guidance (ICL), enabling \texttt{self-navigating} rollouts that improve rollout quality and exploration efficiency.
Finally, \emph{Sample Construction \& Model Optimization} applies \texttt{self-attributing} to convert trajectories and intermediate states into training samples with fine-grained credit assignment (e.g., step-level attribution) and updates the policy parameters accordingly.

\paragraph{Hierarchical Rollout Execution For Agentic Multi-turn Interaction.}
To support complex multi-turn interactions at scale, the rollout process is organized into three hierarchical layers, as illustrated in Figure~\ref{fig:infra_framework}:
\begin{enumerate}
\item \textbf{Service Layer (bottom):} Comprising the \emph{Environment Server} and \emph{LLM Server}, this layer manages multiple independent environment or model workers.
Each worker handles a single environment step or model inference request and exposes standardized interfaces that can be accessed by rollout workers.
\item \textbf{Rollout Workers (middle):} The basic unit for sampling a trajectory.
Each worker interacts with the environment to initialize instances, execute actions, and collect observations, and with the model to assemble prompts and invoke the LLM in sequence, generating a complete trajectory for a given task.
\item \textbf{Rollout Manager / Master (top):} Oversees scheduling of multiple rollout workers, defines termination criteria, and coordinates curriculum strategies across workers.
\end{enumerate}
A key design principle is the \emph{decoupling of agent logic from runtime services}: the environment and LLM are provided as independent services, while the agent logic is encapsulated in a \emph{context manager} (described in Section \ref{sec:cmt}) that composes prompts from the current state.
This separation allows the same rollout infrastructure to host diverse agent workflows—such as ReAct policies, planner–executor schemes, or tool-using agents—without modification to the service layer.

\paragraph{Algorithmic Extensibility.}
The framework is designed to separate algorithmic logic from the underlying infrastructure, so that new methods can be introduced without altering the runtime services:
\begin{itemize}
\item \textbf{Task Manager.} Handles task generation and filtering.
Developers can freely switch between different task–synthesis modules and task–filtering strategies to explore alternative curricula or environment–specific goals.
\item \textbf{Rollout \& Context Manager.} Controls the agent’s multi-turn interaction logic.
By modifying the \emph{context manager}, developers can restructure the agentic flow—for example implementing ReAct, planner–executor, or other reasoning patterns—without changing the service layer.
\item \textbf{Experience Manager.} Summarizes past trajectories into concise experience texts and retrieves them during future rollouts.
Alternative summarization or retrieval mechanisms can be integrated to guide exploration and improve in-context learning.
\item \textbf{Training Pipeline.} Supports interchangeable sample-generation methods and model-optimization algorithms, such as different credit-assignment schemes, advantage estimators, or loss objectives.
\end{itemize}

These modules operate within a unified orchestration loop, allowing researchers to extend or replace any stage while maintaining a stable interface to the rest of the system.

\subsection{Context Manager}
\label{sec:context-manager}

\begin{figure}
  \centering
  \includegraphics[width=1.0\textwidth]{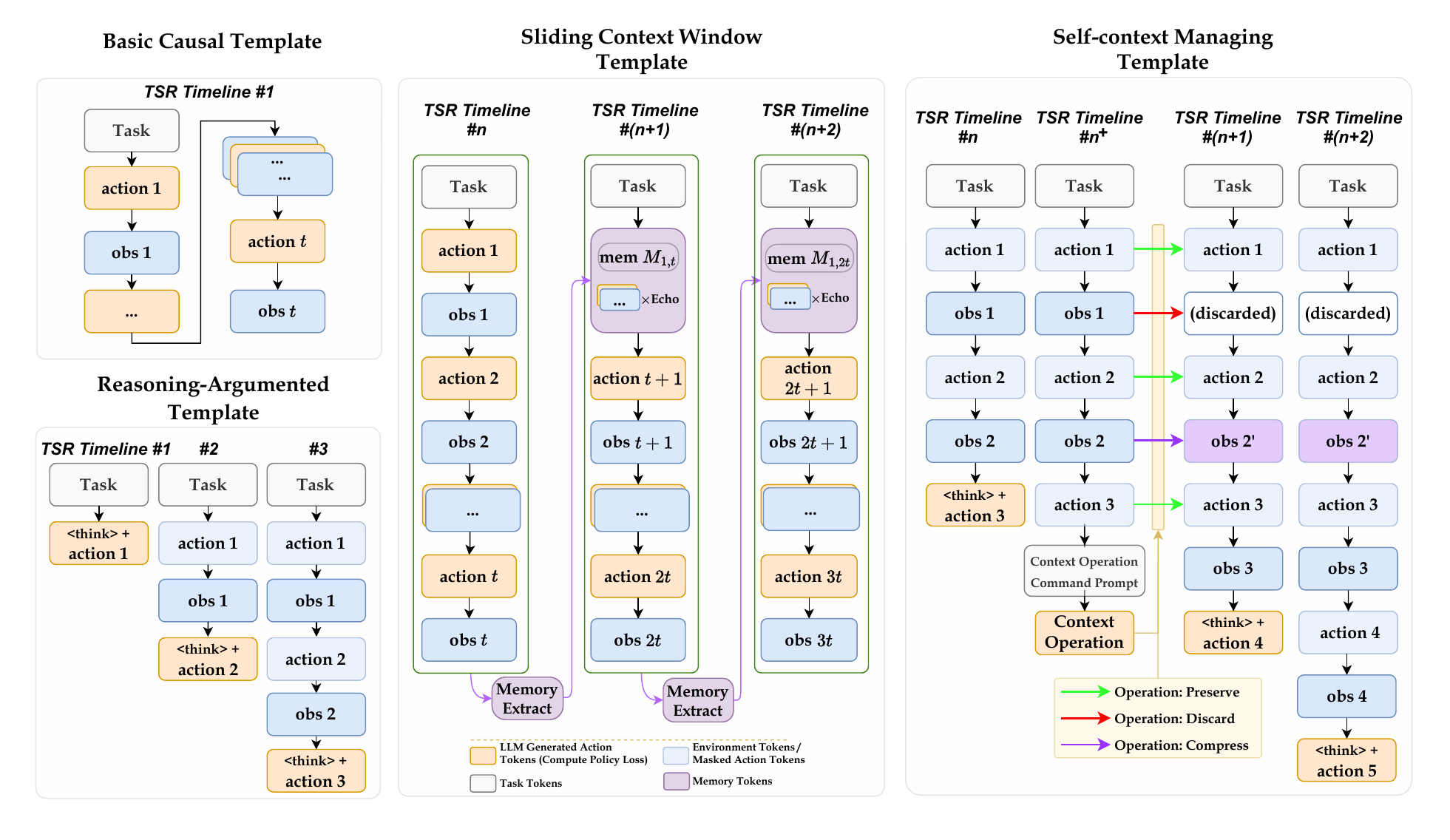}
  \caption{The four fundamental context managing templates of AgentEvolver. Yellow blocks indicate LLM messages, blue blocks correspond to environment messages or messages masked from loss function, and purple blocks represent memory messages. The displayed timelines depict the anticipated outcomes after the Timeline Snapshot Recorder (TSR) completes the timeline merging process.}
  \label{fig:cmts}
\end{figure}

Long-horizon reinforcement learning with large language models (LLMs) requires not only reasoning and planning, but also dynamic management of historical context across multiple interactions. 
Existing paradigms exhibit a trade-off: the \emph{causal multi-step rollout} paradigm maintains strong temporal consistency but lacks flexibility~\citep{jin2025search}, while the \emph{step-independent multi-turn rollout} paradigm offers full editability at the cost of significant computational overhead~\citep{feng2025group}. 
To unify these two perspectives, AgentEvolver introduces a \textbf{Context Manager (CM)} that serves as the central infrastructure for controlling context evolution during agent–environment interactions.

The CM provides a consistent interface for both causal and non-causal interaction modes, enabling efficient, adaptive, and reasoning-aware rollouts. 
Its design aims to (1) maintain the computational efficiency of causal training, (2) allow selective modification of historical context when necessary, and (3) support agents that can eventually \emph{self-manage} their own context. 
To achieve this, CM introduces two fundamental data structures shared across all rollout paradigms.

\paragraph{Core Abstractions: Live Timeline and Snapshot Recorder.}
The Context Manager operates on two simple yet powerful primitives:
\begin{itemize}
    \item \textbf{Live Context Timeline (LCT)} — a mutable sequence representing the agent’s current working context during multi-turn interaction.
    \item \textbf{Timeline Snapshot Recorder (TSR)} — an immutable buffer that stores frozen snapshots of the LCT whenever the policy LLM generates an action.
\end{itemize}

The LCT acts as the agent’s short-term, editable memory, while the TSR maintains a verifiable, token-level record of the entire episode. 
After each rollout, the TSR automatically performs \emph{timeline merging}, removing redundant subsequences and aligning loss masks across overlapping segments. 
This dual-structure design allows CM to support both efficient causal rollouts and flexible reasoning-oriented rollouts within a unified framework.

\subsubsection{Rollout Paradigms and Context-Managing Templates}
\label{sec:cmt}

Built upon the LCT–TSR abstraction, the Context Manager defines a family of \textbf{Context-Managing Templates (CMTs)}. 
Each template specifies how the LCT evolves during interaction, thus representing a different balance between efficiency, reasoning capacity, and autonomy. 
The four fundamental templates are illustrated in Figure~\ref{fig:cmts}.

\paragraph{(i) Basic Causal Template — Efficient and Deterministic.}
The simplest template follows a strictly causal organization of messages. 
At the beginning of an episode, the initial task instruction is appended to the LCT. 
The agent then repeats the observe–act cycle until termination, after which only the final LCT snapshot is retained by the TSR.
This design ensures full temporal consistency and low computational cost, making it ideal for search-style RL training~\citep{jin2025search}. 
However, since messages cannot be edited or removed, memory consumption grows linearly with context length, and the lack of editability complicates reasoning-intensive behaviors.

\paragraph{(ii) Reasoning-Augmented Template — Structured Thinking Before Action.}
Reasoning before action has proven beneficial in complex decision-making tasks~\citep{guo2025deepseek}. 
The reasoning-augmented template trains agents to explicitly \emph{think before acting}. 
For models without pre-trained reasoning ability (e.g., Qwen2~\citep{team2024qwen2}), structured prompts encourage the model to generate \texttt{<think>} tokens before decisions, with additional rewards promoting high-quality reasoning. 
For models that already possess reasoning skills (e.g., Qwen3-14B~\citep{qwen2025technical}), the template enhances consistency and efficiency. 

\paragraph{(iii) Sliding Context Window Template — Scalable to Long Horizons.}
Long-horizon tasks often generate hundreds or thousands of interaction steps, exceeding the token capacity of typical LLMs. 
To address this, the sliding context window template treats the LCT as a moving window. 
When its size exceeds a predefined threshold, the Context Manager summarizes older content into a compressed \textit{memory message} and initializes a new timeline window for subsequent interactions. 
After timeline merging, only a minimal set of window snapshots remains. 
This approach preserves local causality while keeping GPU memory usage constant, achieving scalability to arbitrarily long episodes.

\paragraph{(iv) Self-Context Managing Template — Toward Autonomy.}
The previous templates rely on fixed rules for context updates. 
In contrast, the self-context managing template allows the agent to \emph{actively control} its own memory. 
When the LCT exceeds a token limit, the CM triggers a special \textit{context operation prompt}, asking the policy LLM to select for each message one of three operations:
\[
a_\text{context} \in \{\alpha_\text{keep}, \alpha_\text{remove}, \alpha_\text{compress}\},
\]
corresponding to preservation, deletion, and compression (via an external summarizing LLM).
This paradigm enables fine-grained control over the token budget and encourages the emergence of self-regulated information filtering behavior. 
It is particularly effective in environments that generate large amounts of uninformative text.

\paragraph{Discussion.}
All four templates share the same underlying LCT–TSR mechanism but differ in how they manipulate or delegate control over context evolution. 
Together, they form a spectrum from rigid causality to full autonomy: the Basic Causal Template maximizes efficiency, while the Self-Context Managing Template empowers agents to regulate their own memory dynamically. 
This unified view of context management provides a principled foundation for building scalable, reasoning-capable, and self-adaptive LLM agents.

\subsection{Environment Service: A Scalable and Gym-Compatible Execution Backend}\label{sec:env_service}

To support flexible and high-performance agent-environment interaction, we present an Environment Service that offers a scalable, service-oriented execution backend compatible with the OpenAI Gym interface. In contrast to traditional Gym environments that are tightly coupled with the training loop, our system decouples environment execution into an independent service, enabling remote, concurrent, and isolated environment instances on demand.

\textbf{Standardized Interface} \quad
We implement a minimal, Gym-compatible interface that defines core functionalities such as initialization, state retrieval, step-wise execution, evaluation, and lifecycle management. The interface is fully parameterizable, supporting dynamic runtime configuration of environment behavior. In addition to classical task environments, our design allows seamless integration with open-world tools and modular components such as Model Context Protocols (MCPs) and user-defined functions (UDFs), extending the range of supported scenarios. This unified interface ensures compatibility across diverse experimental settings while maintaining modularity and reusability.

\textbf{High-Concurrency Execution with Ray} \quad
To support large-scale concurrent training, each environment instance is executed as an isolated actor using Ray, a distributed computing framework. This approach provides efficient, lightweight isolation without relying on container-based sandboxing, significantly reducing overhead. Environment instances can be launched and terminated dynamically, enabling flexible scaling across CPU and GPU resources with support for both asynchronous and synchronous interaction modes.

\textbf{Lightweight Deployment and Integration} \quad
The Environment Service is designed for ease of deployment and minimal integration overhead. Preconfigured environments—including \textit{AppWorld} \citep{trivedi2024appworld}, \textit{BFCL} \citep{patil2025bfcl}, \textit{WebShop} \citep{yao2022webshop}, and \textit{Crafter} \citep{hafner2021benchmarking}—can be launched via Docker or installed via Python packages. Once deployed, the service exposes a unified HTTP interface for query submission and agent interaction, allowing training pipelines to operate without concerns over dependency conflicts or environment-specific installations.

\section{Experiments}
\label{sec:experiments}

In this section, we conduct a series of comprehensive experiments to systemically evaluate the performance of \AgentEvolver. We begin by detailing the experimental settings (Section \ref{sec:exp_settings}), followed by a presentation of the main results that demonstrate the superior performance of \AgentEvolver across a broad range of tasks and configurations (Section \ref{sec:main_results}). Finally, we perform extensive ablation and analytical studies to thoroughly investigate the contribution and effectiveness of each module within the proposed framework (Section \ref{sec:experiment_self-questionin} - \ref{sec:experiment_cmt}).

\subsection{Experimental Settings}
\label{sec:exp_settings}
\subsubsection{Benchmarks and Evaluations}
We evaluate \AgentEvolver on two tool-augmented, long-horizon agent benchmarks: \textbf{AppWorld}~\citep{trivedi2024appworld} and \textbf{BFCL~v3} ~\citep{patil2025bfcl}. Both expose multi-step API/tool interactions under sparse terminal rewards and are served through our \emph{Environment Service} (Section \ref{sec:env_service}) with Ray-based isolated actors and a unified HTTP interface to decouple runtime from training. 

For \textbf{AppWorld}, we report \emph{Task Goal Completion (TGC)}---the percentage of tasks for which the agent passes all programmatic evaluation tests---exactly following the official definition. 
For \textbf{BFCL v3}, we \emph{only} use the \emph{multi-turn} split and follow the official multi-turn evaluation: at the end of each turn, an example is marked correct only if it simultaneously passes \emph{state-based} checks (final backend state matches ground truth on non-private attributes) and \emph{response-based} checks (subset-matched execution path); force-terminated runs are counted as incorrect. 
Unless otherwise stated, we report:
(i) \textbf{avg@8}, averaging TGC over 8 independent rollouts per instance; and
(ii) \textbf{best@8} best TGC over 8 independent rollouts evaluation on AppWorld and BFCL v3.
Trajectories are truncated at a maximum of 30 steps.

\subsubsection{Baselines and Backbone Models}
Our experiments leverage two backbone models from the Qwen2.5 series: \textbf{Qwen2.5-7B-Instruct} and \textbf{Qwen2.5-14B-Instruct}~\citep{team2024qwen2} . These models serve as the agent's policy network.
To rigorously evaluate the efficacy of our methods, we compare against a Vanilla GRPO baseline~\citep{deepseek_math}. This baseline follows the standard reinforcement learning paradigm for agents, where the policy is optimized using only the final, sparse outcome signal, without incorporating any of the \AgentEvolver mechanisms such as experience-based guidance or process-based rewards.

\subsubsection{Implementation Details}
We train our agent policies using  \emph{GRPO-style} method.
The general training configuration includes a learning rate of $1 \times 10^{-6}$, a batch size of 32, and 40 epochs per policy update. The KL penalty coefficient is 0.001.
All experiments were conducted on a cluster of 8 NVIDIA A100 (80GB) GPUs using our \AgentEvolver, which is built on PyTorch and the \texttt{veRL} library.

\paragraph{Self-Questioning}
We create environment profiles for the Appworld and BFCL benchmarks, as well as user preferences. The synthetic queries are expected to involve two entities, three attributes, and three operations as defined in environment profiles (e.g. Figure \ref{fig:environment-profile}), and are required to meet a hard difficulty level. For the Appworld environment, exploration steps $N_b$ and $N_d$ are set to $3$ and $17$, respectively. For BFCL, $N_b=3$ and $N_d=27$. We adopt Qwen-Plus as the exploration agent, where the sampling temperature is set to 1. For task synthesis, Qwen-Plus is employed with default settings to achieve balanced efficiency. We use Qwen3-235B-A22B as the LLM judge to better adhere to complex scoring criteria. For task filtering, we first set the lexical deduplication similarity threshold to 0.8 to filter out the queries. During training on hybrid data, we downscale advantages from $p_{\mathrm{train}}$ by a decay factor $0.5$ to reduce bias and stabilize training relative to $p_{\mathrm{target}}$.

\paragraph{Self-Navigating}
For \emph{experience acquisition}, we set $N_{\text{pc}}=4$. The cold-start experience pool is initialized with tasks generated by the \texttt{self-questioning} module in the main experiments. Regarding the experience retrieval, we set $\operatorname{TopK}$ parameter as $k=5$. Implementation details of the construction and retrieval pipelines are provided in ReMe~\footnote{\url{https://github.com/agentscope-ai/ReMe}}.
For \emph{experience-mixed rollout}, the proportion of experience-guided trajectories is set to $\eta = 0.5$. For \emph{experience incorporation}, the importance sampling ratio is clipped at $\epsilon_{\text{low}} = \epsilon_{\text{high}} = 0.28$ with an extended bound $\hat{\epsilon}_{\text{high}} = 0.6$. A comprehensive analysis of sensitivity to these parameters is presented in Section~\ref{sec:experiment_self-navigatin}.
In the ablation study, we report both \emph{avg@4} and \emph{best@4}. Since the experience-mixed rollout ratio is fixed at $0.5$ during testing, we report only the results without experience guidance to provide a fair assessment of the impact of experience incorporation.

\paragraph{Self-Attributing}

We use Qwen-Max~\footnote{\url{https://bailian.console.aliyun.com/}} 
as the judge LLM to perform joint, single-shot evaluation over the \emph{entire} trajectory—i.e., all steps are assessed in one batched call for consistency and efficiency.
Attribution labels are \emph{directly mapped} to signed unit scores (e.g., \texttt{GOOD} $\rightarrow$ 1.0, \texttt{BAD} $\rightarrow$ 0.0 or -1.0)
without intermediate scoring heuristics.
We then apply \textbf{step-level normalization} (within group) to the attribution scores before fusing them with the terminal-outcome signal.
For simplicity, we fix the outcome weight \mbox{$\beta = 1$} and \emph{tune only} the attribution weight $\alpha$ to control the relative contribution of attribution vs.\ outcome; the $\alpha$ schedule and ablation settings are provided in Section ~\ref{sec:experiment_self-attributin}.

Degenerate normalization cases (e.g., near-zero variance) are handled by neutralizing the corresponding signal for stability.

\subsection{Main Results}\label{sec:main_results}

We conduct a series of experiments to evaluate the overall performance of the proposed \AgentEvolver framework and systematically analyze the contributions of its three core components. The results are presented in Table~\ref{tab:main_results}. Following an ablative structure, we begin with a baseline model and progressively incorporate each proposed mechanism —\texttt{self-questioning}, \texttt{self-navigating}, and \texttt{self-attributing}-to isolate and quantify their individual effects.

As shown in Table~\ref{tab:main_results}, integrating all modules into \textbf{AgentEvolver (overall)} leads to substantial performance gains. For the 7B model, overall avg@8 improves by 29.4\% (AppWorld +30.6\%, BFCL v3 +28.1\%) and best@8 by 36.1\% (AppWorld +45.6\%, BFCL v3 +26.6\%). For the 14B model, overall avg@8 increases by 27.8\% (AppWorld +30.7\%, BFCL v3 +24.9\%) and best@8 by 30.3\% (AppWorld +38.0\%, BFCL v3 +22.6\%), demonstrating that AgentEvolver consistently enhances reasoning and task execution across benchmarks.

\begin{table}[t]
\centering
\caption{Performance on two benchmarks. Columns show avg@8 and best@8 for each benchmark, plus their averages (Avg.). All values are in percent (\%). \textbf{Bolded numbers} highlight the best results.
}
\label{tab:main_results}
\small
\setlength{\tabcolsep}{8pt}
\begin{tabular}{@{}ll cc cc cc@{}}
\toprule
\textbf{Model} & \textbf{Params} &
\multicolumn{2}{c}{\textbf{AppWorld}} &
\multicolumn{2}{c}{\textbf{BFCL v3}} &
\multicolumn{2}{c}{\textbf{Avg.}} \\
\cmidrule(lr){3-4}\cmidrule(lr){5-6}\cmidrule(l){7-8}
 & &
avg@8 & best@8 &
avg@8 & best@8 &
avg@8 & best@8 \\
\midrule
Qwen2.5-7B & 7B & 1.8 & 5.6 & 29.8 & 42.4 & 15.8 & 24.0 \\
+Questioning & 7B & 23.2 & 40.3 & 49.0 & 60.6 & 36.1 & 50.5 \\
+Questioning\&Navigating & 7B & 26.3 & 43.1 & 53.3 & 61.0 & 39.8 & 52.1 \\
+Questioning\&Attributing & 7B & 25.7 & 43.7 & 56.8 & 65.3 & 41.3 & 54.5 \\
\rowcolor{gray!20}
\textbf{\AgentEvolver (overall)} & \textbf{7B} &
\textbf{32.4} & \textbf{51.2} &
\textbf{57.9} & \textbf{69.0} &
\textbf{45.2} & \textbf{60.1} \\
\midrule
Qwen2.5-14B & 14B & 18.0 & 31.4 & 41.6 & 54.1 & 29.8 & 42.8 \\
+Questioning & 14B & 44.3 & 65.5 & 60.3 & 72.1 & 52.3 & 68.8 \\
+Questioning\&Navigating & 14B & 45.4 & 65.3 & 62.8 & 74.5 & 54.1 & 69.9 \\
+Questioning\&Attributing & 14B & 47.8 & 65.6 & 64.9 & 76.3 & 56.4 & 71.0 \\
\rowcolor{gray!20}
\textbf{\AgentEvolver (overall)} & \textbf{14B} &
\textbf{48.7} & \textbf{69.4} &
\textbf{66.5} & \textbf{76.7} &
\textbf{57.6} & \textbf{73.1} \\
\bottomrule
\end{tabular}
\end{table}

The sources of this improvement can be further clarified by analyzing the contribution of each component. The most substantial initial gain comes from \textbf{+Questioning}, which enables the agent to autonomously generate diverse training tasks. This mechanism alone raises the avg@8 score from 15.8\% to 36.1\% for the 7B model and from 29.8\% to 52.3\% for the 14B model, effectively mitigating task scarcity and establishing a strong baseline. Building upon this, \texttt{self-navigating} (+Questioning\&Navigating) further enhances exploration efficiency, improving the 7B model’s score to 39.8\% and the 14B model’s to 54.1\%. The addition of \texttt{self-attributing} (+Questioning\&Attributing) brings another notable improvement—to 41.3\% and 56.4\% respectively—demonstrating the benefit of fine-grained credit assignment in optimizing learning efficiency. Finally, the complete \textbf{\AgentEvolver (overall)} achieves the highest performance across all settings, confirming that the three mechanisms—\texttt{self-questioning}, \texttt{self-navigating}, and \texttt{self-attributing}—complement one another synergistically to realize the framework’s full potential.

\subsection{Results of Self-Questioning}\label{sec:experiment_self-questionin}

This section presents detailed evaluations of the proposed \texttt{self-questioning} module, structured into three investigations that validate its contributions. We first compare the synthetic data with the original data from the target task distribution $p_{\mathrm{target}}(g)$ to demonstrate the effectiveness of the self-questioning tasks. Next, we conduct an ablation study on data volume by training the model using varying quantities of synthetic samples, to investigate the impact of data diversity. Finally, we study the LLM Judge, which provides high-quality synthetic signals for model training.

\subsubsection{Effectiveness of Synthetic Data}

\begin{table}[htbp]
    \centering
    \caption{Performance of synthetic data under different data settings. Columns show avg@8 and best@8 for each benchmark, plus their averages (Avg.). All values are in percent (\%).}
    \label{tab:performance-synthetic-data}
    \small
    \begin{tabular}{llcccccc}
    \toprule
         \textbf{Model} & \textbf{Setting} & \multicolumn{2}{c}{\textbf{Appworld}} & \multicolumn{2}{c}{\textbf{BFCL}} & \multicolumn{2}{c}{\textbf{Avg.}} \\ 
         \cmidrule(lr){3-4}\cmidrule(lr){5-6}\cmidrule(l){7-8}
         &  & avg@8 & best@8 & avg@8 & best@8 & avg@8 & best@8  \\ 
    \midrule
        \multirow{4}{*}{Qwen2.5-7B} & Zero-shot & 1.8 & 5.6 & 29.8  & 42.4 & 15.8 & 24.0 \\ 
         & Original $p_\mathrm{target}$  & 16.1 & 25.5 & 58.8 & 74.0 & 37.5  & 49.8  \\ 
         & Synthetic $p_\mathrm{train}$ & 23.2 & 40.3 & 49.0 & 60.6 & 36.1 &  50.5 \\ 
         & Hybrid $p_\mathrm{hybrid}$      & 21.8 & 36.3 & 65.3 & 75.6 & 43.6  & 56.0 \\ 
    \midrule
        \multirow{4}{*}{Qwen2.5-14B} & Zero-shot & 18.0  & 31.4  & 41.6 & 54.1 & 29.8 & 42.8 \\ 
         & Original $p_\mathrm{target}$  & 46.1  & 61.5  & 68.6 & 74.3 & 57.4 & 68.0 \\ 
         & Synthetic $p_\mathrm{train}$ & 44.3  & 65.5  & 60.3 & 72.1 & 52.3 & 68.8 \\ 
         & Hybrid $p_\mathrm{hybrid}$       & 48.4 & 68.1  & 73.0 & 81.1 & 60.7  & 74.6 \\ 
    \bottomrule
    \end{tabular}
\end{table}

To verify the \texttt{self-questioning} module and check whether it can generate effective data to guide training in unknown environments, we first conduct comparative experiments using synthetic data generated by \texttt{self-questioning} module and the original data from the target task distribution. In Table \ref{tab:performance-synthetic-data}, the agents trained on synthetic data in Appworld and BFCL both show significant performance improvements compared to the zero-shot agent, which is also close to the performance achieved with the original data. This suggests that \texttt{self-questioning} explores valuable questions from the environments, and these questions reflect user preferences while being able to guide the agent's capability improvement, achieving results consistent with human-curated data in a more efficient and automatic manner.
Additionally, on the hybrid data, the agent's performance completely surpasses the performance of agents trained on the original data. This suggests that the synthetic data enhances diversity and further expands the boundaries of the original data, thereby extending the agent's capability.

\subsubsection{Influence of Synthetic Data Quantity}

\begin{figure}[tbp]
\centering
\begin{subfigure}[b]{0.40\textwidth}
    \centering
    \includegraphics[width=\textwidth]{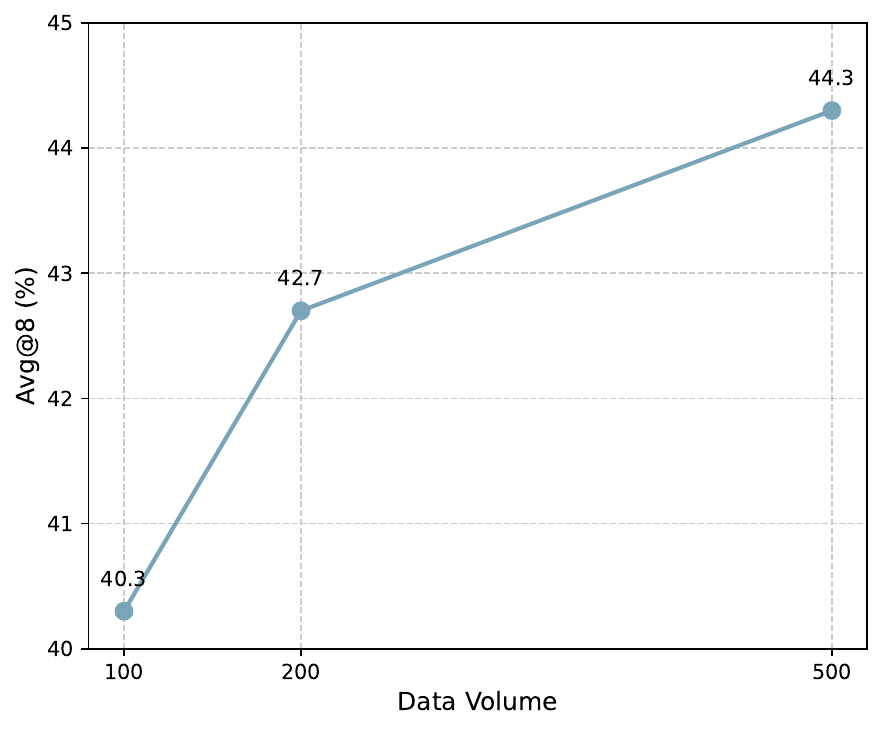}
    \captionsetup{width=.85\linewidth}
    \caption{}
    \label{fig:questioning-performance-vs-data}
\end{subfigure}
\begin{subfigure}[b]{0.33\textwidth}
    \centering
    \includegraphics[width=\textwidth]{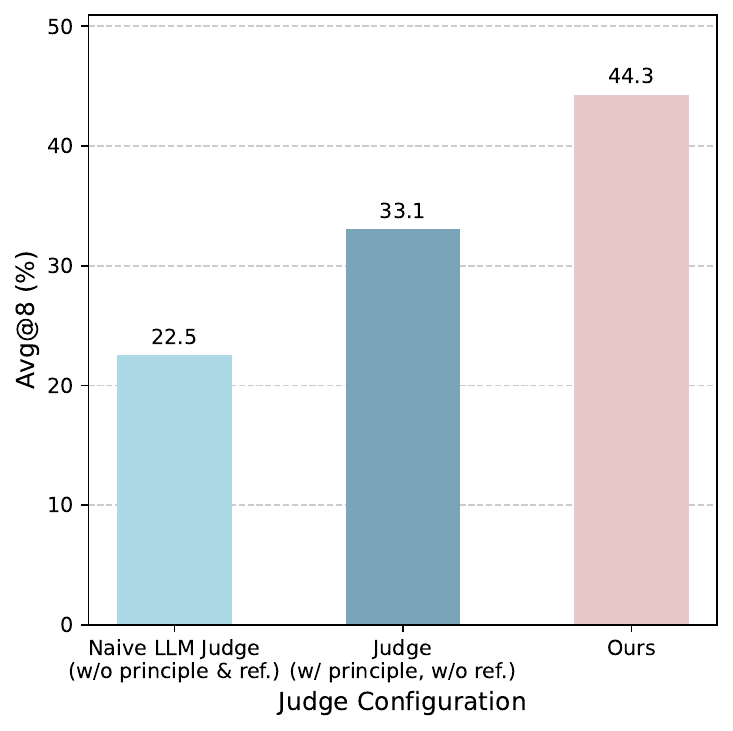}
    \captionsetup{width=.85\linewidth}
    \caption{}
    \label{fig:questioning-performance-vs-judge}
\end{subfigure}
\caption{
(a) Impact of Data Volume on Model Performance. As the volume increases, performance improves accordingly. The diversity of synthetic data effectively enhances performance. (b) Performance Comparison of LLM Judge Configurations. Our proposed judge improves the performance, and the introduction of reference solution significantly enhances it.
}
\end{figure}

Unlike human-curated data, our module efficiently synthesizes a large volume of training data. The underlying assumption is that more data can cover a wider range of scenarios, leading to higher performance. To examine the synthesis efficiency of the \texttt{self-questioning} module, we investigate the influence of data quantity on training. As shown in Figure \ref{fig:questioning-performance-vs-data}, the agent achieves a high performance of 40.3\% with only 100 samples. As the volume increases to 200 and 500, the performance improves further, while the gain diminishes gradually. This suggests that the tasks explored by the \texttt{self-questioning} module are diverse, achieving good training efficiency even with a small amount of data. 
This efficiency stems from the effective exploratory tasks generated by the \texttt{self-questioning} module. Crucially,it reduces the reliance on costly data, opening a path to scale up performance by incorporating large volumes of inexpensive synthetic data.

\subsubsection{Capability Generalization with Synthetic Data}

In this section, we examine the generalization of agent capability trained on synthetic data from a cross-domain perspective. We validate the performance of models trained on Appworld and BFCL synthetic data by cross-evaluating them on each other's test sets, with the results shown in Table \ref{tab:cross-domain-performance-synthetic-data}.
As can be seen, the synthetic data still provides a performance boost for the agent. The Qwen2.5-14B model trained on Appworld and transferred to BFCL exhibits a performance drop of only $4.3\%$. This suggests that the agent's capability is generalizable. The performance improvement brought by our AgentEvolver can be attributed to gains in both general and domain-specific capabilities.
This finding lays a foundation for broader and more generalized agent self-evolution.

\begin{table}[htbp]
    \centering
    \caption{Cross-domain performance of synthetic data. Columns show avg@8 and best@8 for each benchmark, plus their averages (Avg.). \textbf{Bolded numbers} highlight the cross-domain results.}
    \label{tab:cross-domain-performance-synthetic-data}
    \small
    \begin{tabular}{llcccc|cc}
    \toprule
         \textbf{Model} & \textbf{Train on} & \multicolumn{2}{c}{\textbf{Appworld}} & \multicolumn{2}{c}{\textbf{BFCL}} & \multicolumn{2}{|c}{\textbf{Avg.}} \\ 
         \cmidrule(lr){3-4}\cmidrule(lr){5-6}\cmidrule(l){7-8}
         &  & avg@8 & best@8 & avg@8 & best@8 & avg@8 & best@8  \\ 
    \midrule
        \multirow{3}{*}{Qwen2.5-7B} & Zero-shot & 1.8 & 5.6 & 29.8  & 42.4 & 15.8 & 24.0 \\ 
        \cmidrule(lr){2-8}
         & Appworld  & 23.2 & 40.3 & \textbf{36.1} & \textbf{45.0} & 29.7 & 42.7 \\ 
         & BFCL & \textbf{1.2} & \textbf{4.2} & 49.0 & 60.6 & 25.1 & 32.4 \\ 
    \midrule
        \multirow{3}{*}{Qwen2.5-14B} & Zero-shot & 18.0  & 31.4  & 41.6 & 54.1 & 29.8 & 42.8 \\ 
        \cmidrule(lr){2-8}
         & Appworld  & 44.3  & 65.5  & \textbf{56.0} & \textbf{68.9} & 50.2 & 67.2 \\ 
         & BFCL & \textbf{22.9}  & \textbf{40.8}  & 60.3 & 72.1 &  41.6 & 56.5  \\ 
    \bottomrule
    \end{tabular}
\end{table}

\subsubsection{Ablation Study on the LLM Judge}

The LLM Judge is a critical component of the \texttt{self-questioning} module, as it replaces the environment to provide a synthetic reward for the synthetic data. In this part, we conduct experiments to test the efficacy of our proposed LLM Judge, including the scoring process and the reference solution. The relevant results are reported in Figure \ref{fig:questioning-performance-vs-judge}. The figure shows that even if the synthetic data itself is of high quality, the agent's capability can barely be improved if only a naive LLM Judge (w/o principle) is adopted. Performance significantly improves after implementing the \textit{basic principle} proposed in Section \ref{sec:self-questioning-llm-judge}, confirming that the combination of scoring criteria can distinguish the quality of the rollouts. After incorporating the reference solution summarized during the exploration phase, the performance shows substantial improvement, closely matching the performance trained on human-curated data. The reference solution can be specialized to the specific environment to provide a clearer scoring standard while being generalizable across different environments without relying on high-quality human annotation. Through the aforementioned design, our LLM Judge is able to facilitate efficient training in conjunction with the synthetic data.

\subsection{Results of Self-Navigating}\label{sec:experiment_self-navigatin}
In this part, we present a systematic analysis of the \texttt{self-navigating} module along four dimensions. 
First, we validate the \textbf{effectiveness of experience}, demonstrating its ability to guide the model toward higher-quality trajectories. 
Building on this, we compare \textbf{explicit and implicit experience learning}, where the former is realized via ICL and the latter allows the model to internalize optimal behavior patterns without direct exposure to retrieved experiences.
Next, we investigate how the proportion of experience-guided rollouts within the same query group shapes both training and inference, emphasizing the \textbf{balance between exploration and exploitation}. 
Finally, we analyze the influence of $\hat{\epsilon}_{\text{high}}$ in Eq.~\ref{eq:navigating_loss}, revealing its role in balancing \textbf{short- and long-term optimization}.

\begin{figure}[htbp]
\centering
\begin{subfigure}[b]{0.47\textwidth}
    \centering
    \includegraphics[width=\textwidth]{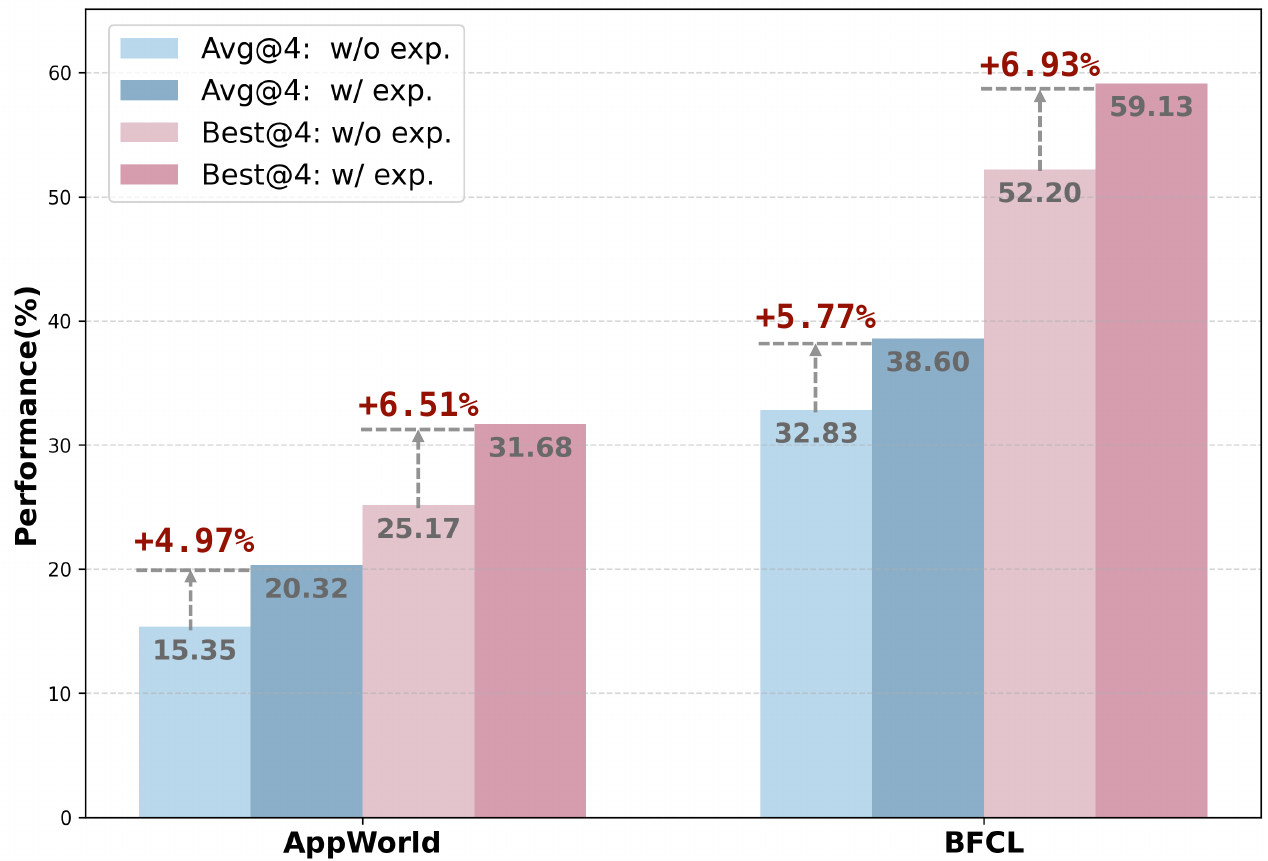}
    \captionsetup{width=.85\linewidth}
    \caption{}
    \label{fig:navi_effect_exp}
\end{subfigure}
\begin{subfigure}[b]{0.49\textwidth}
    \centering
    \includegraphics[width=\textwidth]{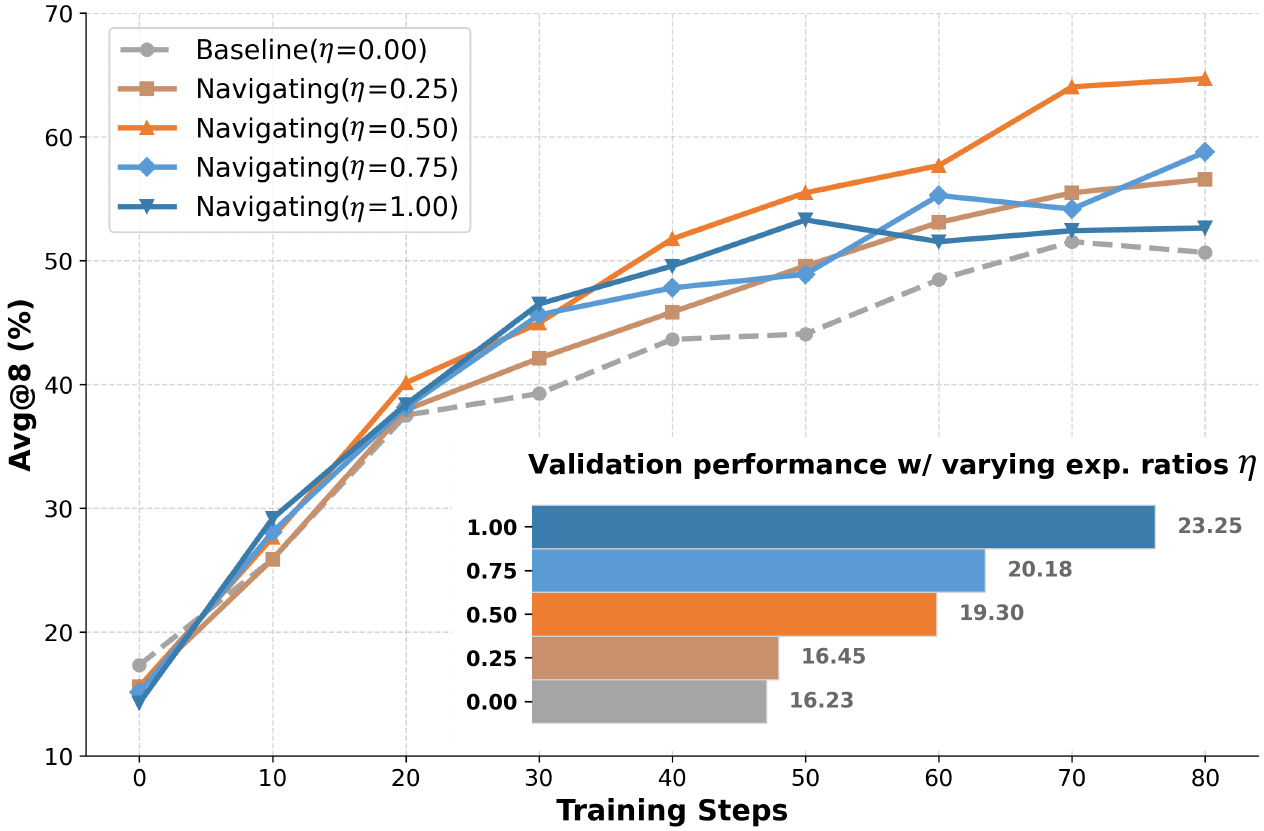}
    \captionsetup{width=.85\linewidth}
    \caption{}
    \label{fig:gamma_analysis}
\end{subfigure}
\caption{
(a) Effect of incorporating experience rollouts during inference. 
(b) Performance trends under varying experience ratios ($\eta$): the lower bars represent zero-shot inference with experience (prior to training), while the upper curve shows training dynamics using experience-guided data, with evaluation performed without experience on the validation set. 
Together, these results reveal the contrast between directly exploiting experience and internalizing it through training.}
\end{figure}

\subsubsection{Effectiveness of Experience on Navigation}
We begin by examining whether incorporating retrieved experiences during inference can improve the quality of generated trajectories without additional training. 
Specifically, we measure trajectory quality by the rewards obtained, which naturally reflect task success. 
As illustrated in Figure~\ref{fig:navi_effect_exp}, results on both AppWorld and BFCL consistently demonstrate that past experiences provide strong navigation signals: they stabilize rollouts and raise the performance ceiling, yielding average gains of \underline{$\uparrow$5.4\% avg@4} and \underline{$\uparrow$6.7\% best@4}.
Collectively, these results underscore the \textbf{value of leveraging experience as an external compass: even without training, contextualized guidance allows the model to explore more effectively and produce superior outcomes.}

\subsubsection{Implicit vs. Explicit Experience Learning}
\begin{table}[htbp]
    \centering
    \caption{Ablation studies of self-navigating on two benchmarks. Columns show avg@4 and best@4 on the \emph{dev} set for each benchmark, plus their averages (Avg.). All values are in percent (\%). \textbf{Bolded numbers} highlight the best results. The backbone model is Qwen2.5-14B. }
    \label{tab:navigate-ablation}
    \small
    \setlength{\tabcolsep}{8pt}
    \begin{tabular}{@{}l cc cc cc@{}}
    \toprule
    \textbf{Method} & 
    \multicolumn{2}{c}{\textbf{AppWorld}} &
    \multicolumn{2}{c}{\textbf{BFCL v3}} &
    \multicolumn{2}{c}{\textbf{Avg.}} \\
    \cmidrule(lr){2-3}\cmidrule(lr){4-5}\cmidrule(lr){6-7}
     & 
    avg@4 & best@4 &
    avg@4 & best@4 &
    avg@4 & best@4 \\
    \midrule
    w/o RL      &  &  &  &  &  & \\
    \quad zero-shot     & 17.3 & 37.4 & 33.0 & 50.6 & 25.2 & 44.0\\
    \quad + \texttt{exp.}    & 20.2 & 41.6 & 41.5 & 61.9 & 30.9 & 51.8\\
    \midrule
    w/ RL      &  &  &  &  &  & \\
    \quad baseline     & 51.5 & 69.8 & 62.8 & 73.0 & 57.2 & 71.4\\
    \quad Navigating (w/o \texttt{select})     & 53.1 & 63.8 & 60.3 & 73.2 & 56.7 & 68.5\\
    \rowcolor{gray!20}
    \quad Navigating (\textbf{ours})     & 64.7 & 85.9 & 65.3 & 73.9 & 65.0 & 79.9\\
    
    \bottomrule
    \end{tabular}
\end{table}

Having established the effectiveness of injecting experience during inference, we next investigate whether such benefits can be sustained and generalized through training. 
Here, we distinguish between \emph{explicit} experience learning, where ICL leverages retrieved trajectories to guide the model, and \emph{implicit} experience learning, where RL-based training enables the policy to internalize experience without retrieval. 
While the explicit approach yields gains, its reliance on external context imposes a clear performance ceiling.
By contrast, implicit learning consistently surpasses both ICL-only (\underline{$\uparrow$34.2\% in avg@4} and \underline{$\uparrow$28.1\% in best@4}) and vanilla RL baselines (\underline{$\uparrow$7.9\% in avg@4} and \underline{$\uparrow$8.5\% in best@4}), achieving superior performance without retrieval at inference (see Table~\ref{tab:navigate-ablation}).
Moreover, ablation studies reveal that removing the \emph{selective boosting} mechanism leads to a clear drop below the RL baseline (\underline{$\downarrow$0.5\% in avg@4} and \underline{$\downarrow$2.9\% in best@4}), confirming its necessity for effectively incorporating experience signals. 
In summary, \textbf{explicit learning yields limited improvements by leveraging external experience, whereas implicit learning internalizes experience to raise the performance ceiling with better generalization and scalability substantially.}

\subsubsection{Exploration vs. Exploitation Dynamics}
Although leveraging retrieved experience enhances trajectory quality through stronger exploitation, the critical question is whether increased exploitation invariably leads to superior outcomes. Greater reliance on prior experience can elevate short-term rewards by guiding rollouts toward higher-quality trajectories; however, RL training fundamentally relies on exploration to preserve trajectory diversity and uncover alternative high-reward solutions. Excessive exploitation thus risks constraining the search space and impeding long-term optimization.
To investigate this trade-off, we vary the proportion $\eta$ of experience-guided rollouts within each query group. 
As expected, inference results show that higher proportions consistently yield greater rewards (lower bars in Figure~\ref{fig:gamma_analysis}), indicating that stronger exploitation directly benefits immediate performance.
However, training dynamics reveal a more nuanced pattern: while larger $\eta$ accelerates early gains, it ultimately suppresses exploration and weakens long-term outcomes, as shown in the upper curve in Figure~\ref{fig:gamma_analysis}.
In contrast, $\eta=0.5$ achieves the most effective balance, enabling the policy to benefit from experience-driven guidance while retaining sufficient exploratory capacity.
Overall, the results indicate that \textbf{while aggressive exploitation benefits inference, sustainable training requires a calibrated balance between exploration and exploitation.}

\begin{figure}
  \centering
  \includegraphics[width=0.85\textwidth]{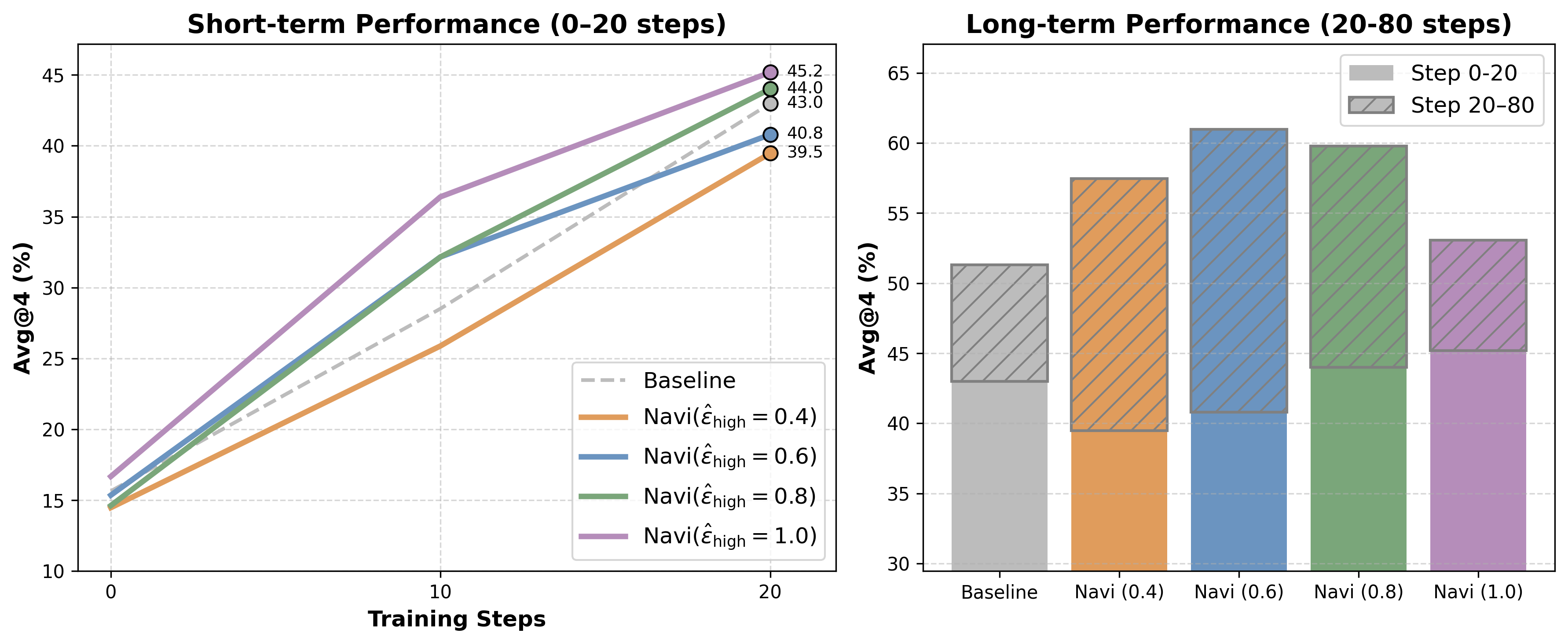}
  \caption{Impact of $\hat{\epsilon}_{\text{high}}$ on learning dynamics. Higher values speed up early learning but harm long-term performance due to overfitting. A value of $\hat{\epsilon}_{\text{high}}=0.6$ offers the best balance between short- and long-term optimization.
  } 
  \label{fig:navi_short-long-term}
\end{figure}

\subsubsection{Short-term vs. Long-term Optimization}
We finally examine how the model balances immediate gains with sustained improvement by studying the effect of the parameter $\hat{\epsilon}_{\text{high}}$ in Eq.~\ref{eq:navigating_loss}.
This parameter relaxes the GRPO clipping threshold, mitigating distribution shifts introduced by experience-guided rollouts and preserving valuable positive samples that might otherwise be discarded.
To assess its impact, we vary $\hat{\epsilon}_{\text{high}}$ across the range $\{0.4, 0.6, 0.8, 1.0\}$. As shown in Figure~\ref{fig:navi_short-long-term}, we find that larger values accelerate early learning (within the first 20 steps) but tend to compromise long-term performance (up to 80 steps) due to overfitting to biased trajectories. Our findings indicate that setting $\hat{\epsilon}_{\text{high}}=0.6$ achieves the most favorable balance, facilitating both rapid initial progress and stable long-term optimization.
These observations suggest that \textbf{moderate relaxation fosters early progress while maintaining robustness, whereas overly aggressive values may deliver short-term gains at the cost of long-term generalization.}

\subsection{Results of Self-Attributing}
\label{sec:experiment_self-attributin}
This section evaluates the \texttt{self-attributing} mechanism proposed in Section~\ref{sec:self-attribution}, which assigns fine-grained rewards to trajectory steps based on their contribution. We examine three aspects: effectiveness through ablation studies comparing outcome-only, attribution-only, and dual-channel configurations (Section~\ref{sec:attr_effectiveness}); sample efficiency by analyzing convergence speed and training dynamics (Section~\ref{sec:attr_efficiency}); and the impact of the attribution weight hyperparameter $\alpha$ on learning trajectories (Section~\ref{sec:attr_hyperparameter}). Following the main experimental setup, all models are trained exclusively on synthetic data generated by \texttt{self-questioning}, using Qwen2.5-7B and Qwen2.5-14B as backbone models on AppWorld and BFCL v3 benchmarks.

\subsubsection{Effectiveness of Self-Attributing} 
\label{sec:attr_effectiveness}
To validate the effectiveness of our \texttt{self-attributing} mechanism (Section~\ref{sec:self-attribution}), we conduct systematic ablation studies comparing four configurations on AppWorld and BFCL v3: (i) Backbone, the Qwen2.5 models (7B and 14B) without any reinforcement learning; (ii) \textbf{Attributing}, our complete \texttt{self-attributing} method that integrates both attribution rewards $r^{\text{attr}}$ and outcome rewards $r^{\text{out}}$ with $\alpha{=}0.1$ to balance their contributions; (iii) \textbf{w/o $r^{\text{attr}}$}, an ablation that removes step-wise attribution signals and relies solely on terminal outcome rewards; and (iv) \textbf{w/o $r^{\text{out}}$}, an ablation that excludes terminal rewards and depends exclusively on attribution-based feedback. All variants are evaluated using avg@8 and best@8 metrics under the TGC evaluation protocol (Table~\ref{tab:attributing_Effectiveness}).  Note that AppWorld results are reported on the dev set here, while Table~\ref{tab:main_results} uses the test-normal set.

\begin{table}[h]
\centering
\caption{Effectiveness of \texttt{self-attributing}. Columns show avg@8 and best@8 on the \emph{dev} set for each benchmark, plus their averages (Avg.). All values are in percent (\%). \textbf{Bolded numbers} highlight the best results.}
\label{tab:attributing_Effectiveness}
\small
\setlength{\tabcolsep}{8pt}
\begin{tabular}{@{}ll cc cc cc@{}}
\toprule
\textbf{Model} & \textbf{Params} &
\multicolumn{2}{c}{\textbf{AppWorld}} &
\multicolumn{2}{c}{\textbf{BFCL v3}} &
\multicolumn{2}{c}{\textbf{Avg.}} \\
\cmidrule(lr){3-4}\cmidrule(lr){5-6}\cmidrule(l){7-8}
 & &
avg@8 & best@8  &
avg@8  & best@8  &
avg@8  & best@8  \\
\midrule
Qwen2.5-7B   & 7B   & 3.1 & 9.1 & 29.8 & 42.4 & 16.4 & 25.7  \\
\rowcolor{gray!20}
\textbf{Attributing } & \textbf{7B} & \textbf{38.4} & \textbf{57.1} & \textbf{56.8} & \textbf{65.3} & \textbf{47.6} & \textbf{61.2} \\
 - w/o $\hat{r}^{\text{attr}}$  & 7B & 33.6 & 46.6 & 49.0 & 60.6 & 41.3 & {53.6} \\
 - w/o $\hat{r}^{\text{out}}$ & 7B & 20.2 & 37.5 & 51.3 & 65.2 & {35.7} & {51.4} \\
\midrule
Qwen2.5-14B   & 14B   & 17.8 & 27.7 & 41.8 & 55.3 & 29.8 & 41.5  \\
\rowcolor{gray!20}
\textbf{Attributing} & \textbf{14B} & \textbf{59.2} & \textbf{75.1} & \textbf{64.9} & \textbf{76.3} & \textbf{62.0} & \textbf{75.7} \\
 - w/o $\hat{r}^{\text{attr}}$ & 14B & 54.6 & 71.3 & 60.3 & 72.1 & 57.4 & 71.7 \\
 - w/o $\hat{r}^{\text{out}}$ & 14B &  42.5 & 60.3 & 63.4 & 73.5 & {53.0} & {66.9} \\

\bottomrule
\end{tabular}
\end{table}

Our \texttt{self-attributing} method (Attributing) achieves the best performance across all settings, demonstrating substantial improvements over the baseline. On the 7B model, it attains a macro-average of 47.6\% avg@8, compared to {Qwen2.5}'s 16.4\% and 25.7\% respectively. The 14B model shows even stronger results, reaching 62.0\% avg@8 and 75.7\% best@8, significantly outperforming the baseline's 29.8\% and 41.5\%. On AppWorld specifically, \texttt{self-attributing} improves avg@8 from 3.1\% to 38.4\% on the 7B model and from 17.8\% to 59.2\% on the 14B model, representing dramatic gains of 35.3\% and 41.4\% percentage points respectively.

The ablation studies reveal the complementary necessity of both reward channels. The {w/o $\hat{r}^{\text{attr}}$} variant, relying solely on outcome rewards, achieves reasonable performance but suffers from uniform credit assignment across trajectory steps. More notably, the {w/o $\hat{r}^{\text{out}}$} variant shows that process-level signals alone are insufficient: while it substantially improves over {Qwen2.5}, it consistently underperforms {w/o $\hat{r}^{\text{attr}}$} across both model scales. This indicates that attribution provides valuable intermediate guidance but requires terminal outcomes to ground policy learning toward task objectives. The consistent gains of our complete method validate the dual-channel design: combining independently normalized signals for intermediate step quality ($\hat{r}^{\text{attr}}$) and final outcome effectiveness ($\hat{r}^{\text{out}}$) enables the policy to optimize both the reasoning process and task success.

\subsubsection{Sample Efficiency of Self-Attributing}
\label{sec:attr_efficiency}
\begin{figure}[h!]
\centering
\begin{subfigure}[b]{0.48\textwidth}
    \centering
    \includegraphics[width=\textwidth]{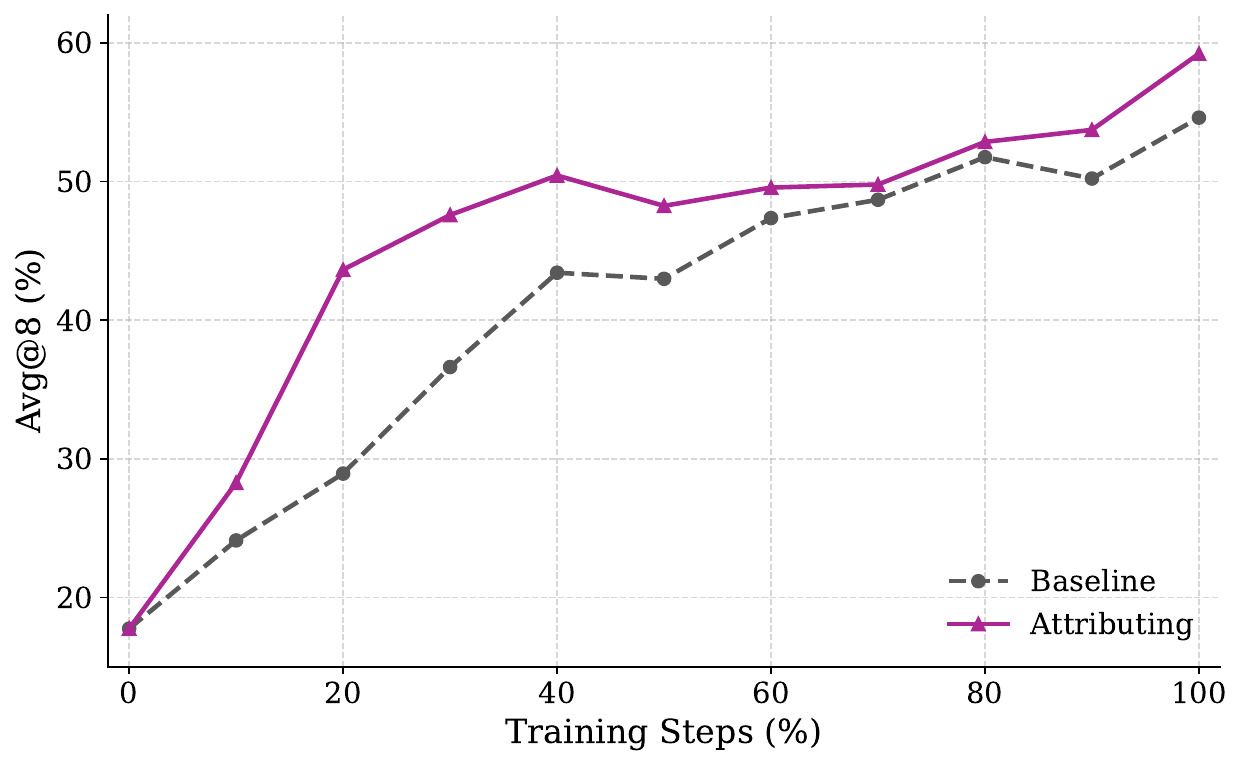}
    \caption{AppWorld}
    \label{fig:appworld}
\end{subfigure}
\begin{subfigure}[b]{0.48\textwidth}
    \centering
    \includegraphics[width=\textwidth]{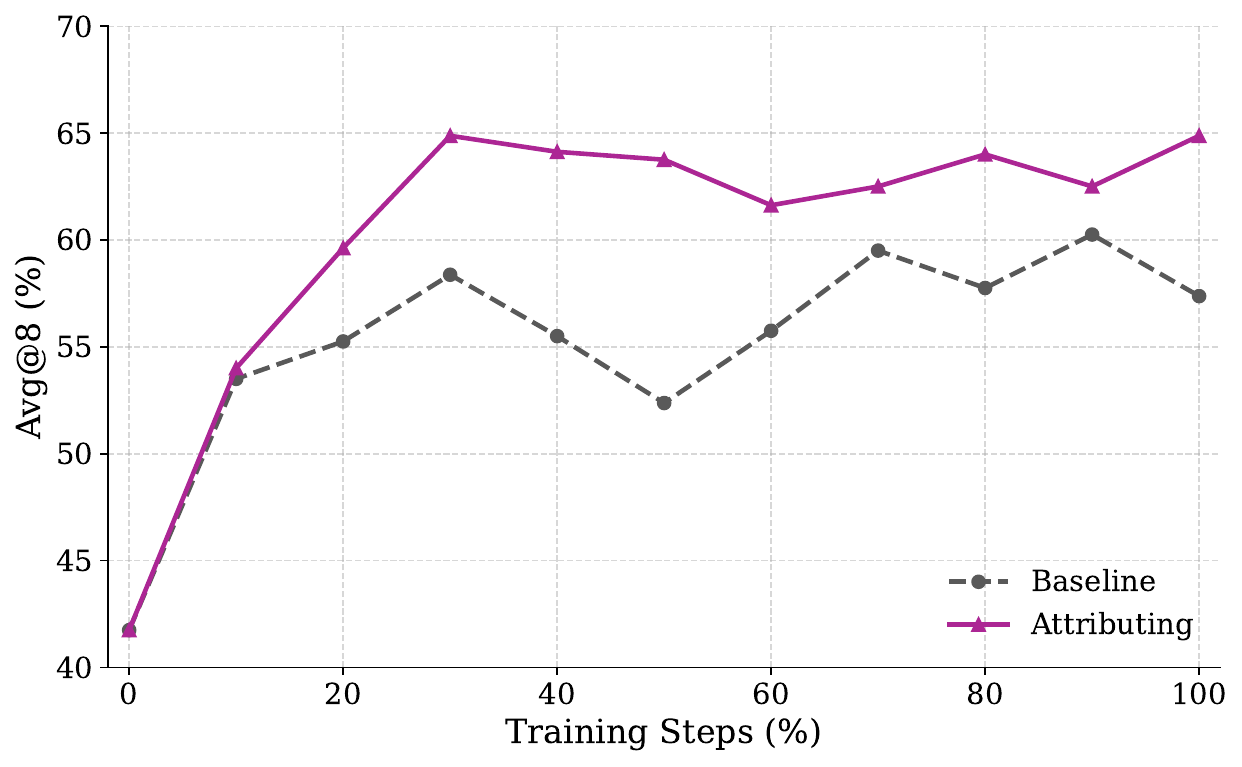}
    \caption{BFCL v3}
    \label{fig:bfclv3}
\end{subfigure}
\caption{\textbf{Efficiency under equal training budgets.} Learning curves on the validation sets of the (a) AppWorld and (b) BFCL v3 benchmarks. Our method, Attributing (solid purple line), is compared against the GRPO baseline (dashed grey line).}
\label{fig:attr_sample_efficiency} 
\end{figure}

\begin{figure}[t!]
  \begin{minipage}{0.51\linewidth}
    \centering
    \includegraphics[width=0.98\textwidth]{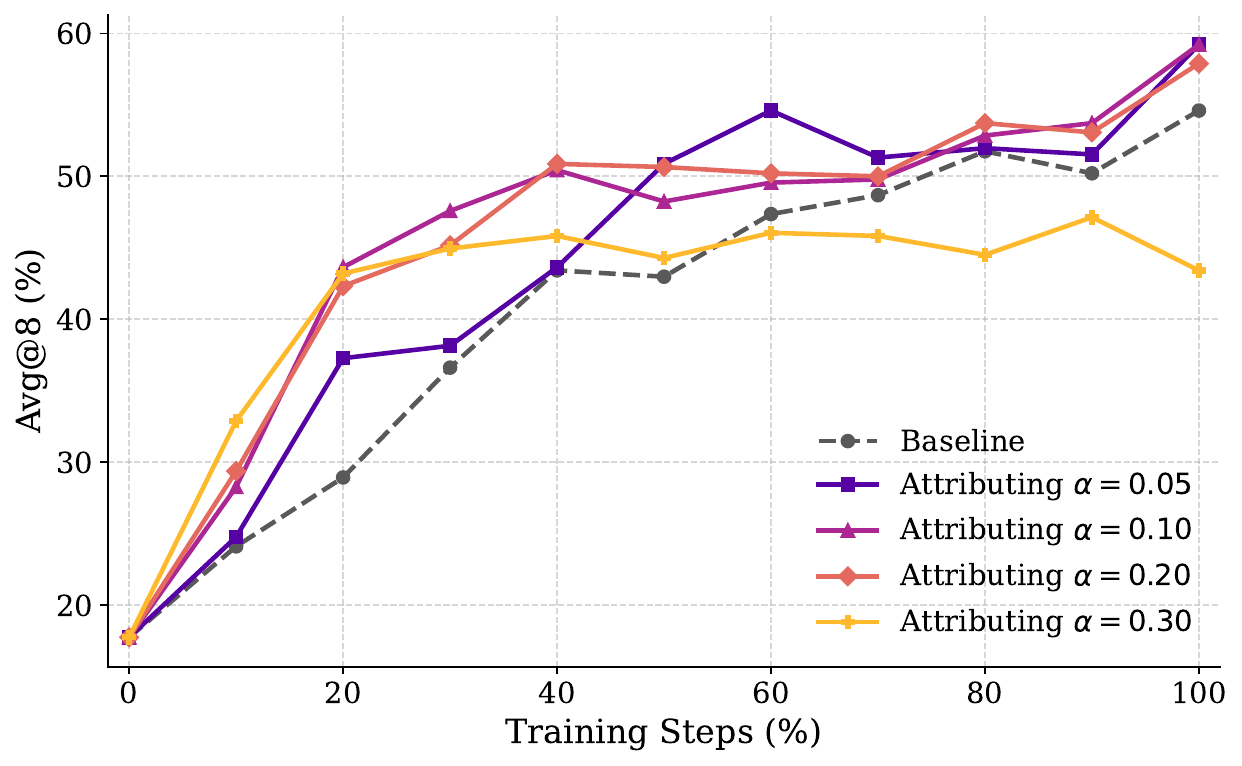}
    \vspace{-6pt}
    \caption{\textbf{Learning curves for different $\alpha$ values} on AppWorld. The dashed line shows the outcome-only baseline; colored lines correspond to Attributing with $\alpha \in \{0.05, 0.10, 0.20, 0.30\}$ (darker to lighter).}
    \label{fig:attr_alpha_curves}
  \end{minipage}
  \hspace{2pt}
\begin{minipage}{0.43\linewidth}
    \centering
    \captionof{table}{\textbf{Sample efficiency metrics} for the Qwen2.5-14B backbone model. Steps@.9 measures the training steps required to reach 90\% of the baseline's best performance (lower is better, ↓).}
    \label{tab:attr_efficiency_metrics}
    \footnotesize
    \begin{threeparttable}
      \setlength{\tabcolsep}{6pt} 
      \renewcommand{\arraystretch}{1.05}
      \begin{tabular}{l cc} 
        \toprule
        \textbf{Model} & \makecell{\textbf{AppWorld} \\ Steps@.9 (↓)} & \makecell{\textbf{BFCL v3} \\ Steps@.9 (↓)} \\
        \midrule
        Baseline & 90 & 60 \\
        \rowcolor{gray!15}
        \textbf{Attributing} & \textbf{40} & \textbf{20} \\
        \bottomrule
      \end{tabular}
      \vspace{2pt}
    \end{threeparttable}
\end{minipage}
  \vspace{-8pt}
\end{figure}

Beyond improving final performance (Section~\ref{sec:self-attribution}), we now investigate whether the \texttt{self-attributing} mechanism also enhances sample efficiency—a critical consideration for practical agent training. To this end, we conduct a quantitative analysis from two complementary perspectives: convergence speed, measured by the training steps needed to reach a performance threshold, and overall learning throughput, quantified by the Area Under the Curve (AUC). The learning trajectories are visualized in Figure~\ref{fig:attr_sample_efficiency}, with corresponding metrics for the Qwen2.5-14B model detailed in Table~\ref{tab:attr_efficiency_metrics}.

Our Attributing method demonstrates substantial improvements in both convergence speed and overall training efficiency. To reach 90\% of the baseline's final performance, our method requires only 40 training steps on AppWorld (baseline: 90 steps) and 20 steps on BFCL v3 (baseline: 60 steps), corresponding to reductions of 55\% and 67\% respectively. The method also achieves higher area-under-curve scores: 46.26 compared to 41.03 on AppWorld, and 61.02 compared to 55.78 on BFCL v3. These trends are consistent across both benchmarks, as shown in the learning curves of Figure~\ref{fig:attr_sample_efficiency}.

This efficiency gain arises from fine-grained credit assignment. Step-wise attribution signals $\hat{r}^{\text{attr}}$ provide dense feedback at each decision point, complementing sparse terminal outcomes $\hat{r}^{\text{out}}$ in the composite reward. Each trajectory thus yields richer gradient information, reducing variance and minimizing redundant exploration.

\subsubsection{Hyperparameter Analysis}
\label{sec:attr_hyperparameter}
To understand the trade-off between rapid initial convergence and robust long-term performance, we analyze the hyperparameter $\alpha$ from our composite reward $r^{\text{comp}}$ (Eq.~23). This parameter controls the relative influence of the process-oriented guidance from attribution signals ($\hat{r}^{\text{attr}}$) compared to the objective-driven supervision from terminal outcomes. We systematically evaluate several values for $\alpha \in \{0.05, 0.10, 0.20, 0.30\}$, with the resulting learning curves for the Qwen2.5-14B backbone on AppWorld shown in Figure~\ref{fig:attr_alpha_curves}.

The results reveal a trade-off between early convergence and long-term performance. In the early stage of training, larger $\alpha$ values accelerate learning: $\alpha{=}0.30$ (yellow line) reaches 45\% at 20 steps, while the baseline achieves only 28\%. However, this early advantage does not persist. By the end of training, $\alpha{=}0.30$ degrades to 43\%, falling below all other attribution-based settings. In contrast, $\alpha{=}0.10$ (purple line) and $\alpha{=}0.20$ (pink line) maintain strong performance throughout, converging to approximately 59\%—substantially higher than both the baseline (55\%) and the high-$\alpha$ variant. The setting $\alpha{=}0.05$ (dark purple line) shows more conservative early gains but achieves comparable final performance.

This pattern indicates that excessive reliance on attribution signals can harm long-term learning. Large $\alpha$ provides strong process-level guidance that accelerates initial policy improvement, but may cause overfitting to the LLM judge's heuristic labels rather than true task objectives. Conversely, small $\alpha$ underutilizes the attribution channel, slowing early progress. The optimal range $\alpha{\in}[0.10, 0.20]$ balances exploitation of dense attribution feedback with grounding in outcome-based supervision, yielding both rapid convergence and robust final performance.

\subsection{Effectiveness of Context-Managing Templates}\label{sec:experiment_cmt}

Different context-management strategies substantially influence an agent’s ability to adapt to diverse tasks. 
In this section, we evaluate the effectiveness of the four fundamental Context-Managing Templates (CMTs) introduced in Section~\ref{sec:context-manager}, using the \textit{AppWorld} benchmark and evaluating agent's TGC and Scenario Goal Completion (SGC) metrics. 
All experiments are conducted with the Qwen3-14B model as the underlying policy backbone.

\begin{table}[htbp]
\centering
\caption{
\textbf{Comparison of TGC@4, TGC@8, SGC@4, and SGC@8 metrics across different Context-Managing Templates (CMTs) in the AppWorld benchmark.} 
The base model is Qwen3-14B. 
Higher values indicate better performance (↑).
}
\label{tab:results-cmt}
\small
\setlength{\tabcolsep}{7pt}
\begin{tabular}{@{}l cc cc@{}}
\toprule
\textbf{Context-Managing Template (CMT)} &
\multicolumn{2}{c}{\textbf{TGC}} &
\multicolumn{2}{c}{\textbf{SGC}} \\
\cmidrule(lr){2-3}\cmidrule(lr){4-5}
 & @4 (↑) & @8 (↑) & @4 (↑) & @8 (↑) \\
\midrule
Basic Causal Template                      & 0.435          & 0.506             & 0.268             & 0.375 \\
Reasoning-Augmented Template               & \textbf{0.661} & 0.690             & \textbf{0.500}    & \textbf{0.571} \\
Sliding-Window Template                    & 0.560          & 0.601             & 0.393             & 0.411 \\
Self-Context-Managing Template             & 0.613          & \textbf{0.720}    & \textbf{0.500}    & \textbf{0.607} \\
\bottomrule
\end{tabular}
\end{table}

As shown in Table~\ref{tab:results-cmt}, the \textbf{Self-Context-Managing Template (SCMT)} achieves the best overall performance in the AppWorld benchmark, particularly on the long-horizon metric \textbf{TGC@8}. 
AppWorld requires agents to reason over extensive lists of available tool APIs, where only a small subset is relevant to the current task. 
The SCMT improves efficiency by dynamically compressing or discarding redundant context segments, enabling more focused reasoning and streamlined agent–environment interactions.

The \textbf{Reasoning-Augmented Template (RAT)} ranks second, highlighting the benefit of explicit ``think-before-act'' steps in enhancing the reasoning ability of Qwen3 within complex multi-tool environments. 
The \textbf{Sliding-Window Template (SWT)} performs moderately, as its periodic memory summarization can omit essential dependencies over longer sequences, reducing contextual consistency compared to SCMT. 
Finally, the \textbf{Basic Causal Template (BCT)} provides a simple yet robust baseline but lacks adaptive memory control, leading to lower task completion scores in extended contexts.

\section{Conclusion and Next Steps}
\label{sec:conclusion}
This work introduced \textbf{AgentEvolver}, a self-evolving agent framework that shifts the learning initiative from fixed, human-engineered pipelines to LLM-guided improvement. Grounded in a standard training flow---from environments to tasks, from tasks to trajectories, and from trajectories to policy---AgentEvolver operationalizes three synergistic mechanisms:
(i) \texttt{self-questioning} for curiosity-driven task generation,
(ii) \texttt{self-navigating} for experience-guided exploration, and
(iii) \texttt{self-attributing} for fine-grained credit assignment along long horizons.
Together, these components systematically alleviate three persistent bottlenecks in RL-driven agent training: task scarcity, inefficient exploration, and low sample utilization. 

Beyond methodology, we presented a practical infrastructure that standardizes environment interfaces, integrates with RL stacks such as \texttt{veRL}, and supports modular extension. This combination of algorithmic design and engineering pragmatics positions AgentEvolver as both a research vehicle and a reusable foundation for building adaptive, tool-augmented agents.

\subsection*{Next Steps}
Looking ahead, we outline three directions that we believe will most effectively extend the scope and impact of AgentEvolver:

\paragraph{1.\ Challenge-oriented applications.}
Building on current validations in \textsc{AppWorld} and \textsc{BFCL}, we will target environments with higher task complexity and greater real-world value (e.g., multi-API enterprise workflows, safety-critical tool chains, and long-horizon, interleaved objectives). The goal is to \emph{co-design} environments and self-questioning curricula so that the agent uncovers functional boundaries more systematically and produces deployable models with measurable business or scientific impact.

\paragraph{2.\ Scaling to larger models.}
We will explore AgentEvolver with larger LLM-based policy models to investigate how the benefits of self-evolution scale with model capacity. Key questions include: How does task quality under \texttt{self-questioning} improve as reasoning depth and reflective ability increase with scale? Do larger policies, capable of abstracting and reusing their prior experiences, exhibit reduced exploration redundancy during \texttt{self-navigating}? Can \texttt{self-attributing} achieve more accurate credit assignment when enhanced causal reasoning enables finer-grained tracing of decision origins? We will further examine compute–data trade-offs and develop cost-aware curricula to enable scalable and efficient long-horizon training.

\paragraph{3.\ LLM-level self-evolving.}
We aim to unify exploration, experience abstraction, self-attribution, and inference within a single model (or tightly coupled family of models), enabling \emph{LLM-level self-improvement}. Concretely, the same model will (i) generate tasks and hypotheses, (ii) navigate new environments with guidance from distilled priors, (iii) attribute outcomes to internal decisions, and (iv) update its reasoning patterns and control policies accordingly. This line pursues tighter loops between memory, generalization, and policy improvement, potentially yielding agents that maintain and refine a self-consistent body of procedural knowledge over time.

\medskip
Pursuing these directions will stress-test the proposed mechanisms in more realistic and dynamically evolving settings. By coupling larger-capacity models with end-to-end, LLM-level self-evolving loops, we expect further gains in sample efficiency, robustness, and cross-environment generalization, bringing us closer to scalable, cost-effective, and continually improving agentic systems.





\bibliography{colm2025_conference}
\bibliographystyle{colm2025_conference}


\end{document}